%% file: main.tex
\documentclass{article}
\usepackage[numbers]{natbib}

\usepackage[preprint]{neurips_2025}

\usepackage[utf8]{inputenc} 
\usepackage[T1]{fontenc}    
\usepackage{hyperref}       
\usepackage{url}            
\usepackage{booktabs}       
\usepackage{amsfonts}       
\usepackage{nicefrac}       
\usepackage{microtype}      
\usepackage{xcolor}         

\usepackage{times}
\usepackage{epsfig}
\usepackage{amsmath}

\usepackage{array}
\usepackage{subcaption}
\usepackage{multirow,multicol}
\usepackage[flushleft]{threeparttable}
\usepackage{comment}
\usepackage{algpseudocode}
\usepackage{algorithm}
\usepackage{booktabs}
\usepackage{bbm, dsfont}
\usepackage[font=small,labelfont=bf]{caption}

\usepackage{amssymb}
\usepackage{pifont}

\newcommand{\app}{\raise.17ex\hbox{$\scriptstyle\sim$}}

\newlength\savewidth

\makeatletter\renewcommand\paragraph{\@startsection{paragraph}{4}{\z@}
  {.5em \@plus1ex \@minus.2ex}{-.5em}{\normalfont\normalsize\bfseries}}\makeatother

\def\tablecite#1#{%
  \def\pretablecite{#1}%
  \tableciteaux}
\def\tableciteaux#1{%
  \textsuperscript{\expandafter\originalcite\pretablecite{#1}}%
}

\usepackage{graphicx}
\usepackage{enumitem}
\usepackage{wrapfig}
\usepackage{lipsum}

\newcolumntype{H}{>{\setbox0=\hbox\bgroup}c<{\egroup}@{}}
\newcolumntype{a}{>{\columncolor{Gray}}c}

\usepackage{tabu}
\usepackage{nicematrix}

\definecolor{ForestGreen}{rgb}{0.13, 0.55, 0.13}
\definecolor{Green}{rgb}{0.0, 0.5, 0.0}
\definecolor{green(munsell)}{rgb}{0.0, 0.66, 0.47}
\definecolor{green(ryb)}{rgb}{0.4, 0.69, 0.2}
\definecolor{green(pigment)}{rgb}{0.0, 0.65, 0.31}
\definecolor{citecolor}{HTML}{0071bc}
\definecolor{GrayXMark}{gray}{0.7}

\usepackage{tabularx}
\usepackage[export]{adjustbox}

\definecolor{ForestGreen}{rgb}{0.13, 0.55, 0.13}
\definecolor{Green}{rgb}{0.0, 0.5, 0.0}
\definecolor{green(munsell)}{rgb}{0.0, 0.66, 0.47}
\definecolor{green(ryb)}{rgb}{0.4, 0.69, 0.2}
\definecolor{green(pigment)}{rgb}{0.0, 0.65, 0.31}

\newcolumntype{x}[1]{>{\centering\let\newline\\\arraybackslash\hspace{0pt}}p{#1}}
\definecolor{Gray}{gray}{0.9}

\usepackage{makecell}

\definecolor{ForestGreen}{rgb}{0.13, 0.55, 0.13}
\definecolor{Green}{rgb}{0.0, 0.5, 0.0}
\definecolor{green(munsell)}{rgb}{0.0, 0.66, 0.47}
\definecolor{green(ryb)}{rgb}{0.4, 0.69, 0.2}
\definecolor{green(pigment)}{rgb}{0.0, 0.65, 0.31}
\definecolor{citecolor}{HTML}{0071bc}
\definecolor{GrayXMark}{gray}{0.7}

\usepackage{tabularx}
\usepackage[export]{adjustbox}

\usepackage{hyperref}
\hypersetup{
    colorlinks,
    linkcolor={black},
    citecolor={black},
    urlcolor={black}
}
\usepackage[capitalize]{cleveref}
\crefname{section}{Sec.}{Secs.}
\Crefname{section}{Section}{Sections}
\Crefname{table}{Table}{Tables}
\crefname{table}{Table}{Tabs.}

\definecolor{ForestGreen}{rgb}{0.13, 0.55, 0.13}
\definecolor{Green}{rgb}{0.0, 0.5, 0.0}
\definecolor{green(munsell)}{rgb}{0.0, 0.66, 0.47}
\definecolor{green(ryb)}{rgb}{0.4, 0.69, 0.2}
\definecolor{green(pigment)}{rgb}{0.0, 0.65, 0.31}

\usepackage{colortbl}
\newcommand{\cmark}{\text{\ding{51}}}%
\newcommand{\xmark}{\text{\ding{55}}}%
\usepackage{xspace}

\usepackage{graphicx}
\usepackage{wrapfig}

\usepackage{mathtools}
\DeclarePairedDelimiterX\set[1]\lbrace\rbrace{#1}

\usepackage{pgfplots}
\usepackage{tikz}
\usepackage{enumitem}

\usepackage{tikzsymbols}

\newcommand{\ours}{LIFT\xspace}

\title{Language-Image Alignment with Fixed Text Encoders}

\author{
\begin{tabular}{c}
Jingfeng Yang$^{1*}$ \quad Ziyang Wu$^{1*}$ \quad Yue Zhao$^1$ \quad Yi Ma$^{1,2}$
\end{tabular} \vspace{4pt} \\
$^1$UC Berkeley \quad $^2$The University of Hong Kong \\
}

\begin{document}
\maketitle

\vspace{-2mm}
\vspace{-3pt}
\begin{abstract} 

Currently, the most dominant approach to establishing language-image alignment is to pre-train text and image encoders jointly through contrastive learning, such as CLIP and its variants. In this work, we question whether such a costly joint training is necessary. In particular, we investigate if a pre-trained fixed large language model (LLM) offers a good enough text encoder to guide visual representation learning. That is, we propose to learn Language-Image alignment with a Fixed Text encoder (LIFT) from an LLM by training only the image encoder. Somewhat surprisingly, through comprehensive benchmarking and ablation studies, we find that this much simplified framework LIFT is highly effective and it outperforms CLIP in most scenarios that involve compositional understanding and long captions, while achieving considerable gains in computational efficiency. Our work takes a first step towards systematically exploring how text embeddings from LLMs can guide visual learning and suggests an alternative design choice for learning language-aligned visual representations. Our code and checkpoints are available at \url{https://github.com/Jingfeng0705/LIFT}.

\end{abstract}
\vspace{-3pt}
\input{sections/1_introduction}
\input{sections/2_related_works}

\input{sections/3_method}
\input{sections/4_experiements}

\input{sections/5_conclusion}



\newpage
{\small
\bibliographystyle{abbrv}
\bibliography{references}
}

\newpage
\input{sections/6_appendix}

\end{document}

%% file: sections/1_introduction.tex
\def\IntroDemo#1{
    \captionsetup[sub]{font=small}
    \begin{figure*}[#1]
      \centering
      \includegraphics[width=0.99\linewidth]{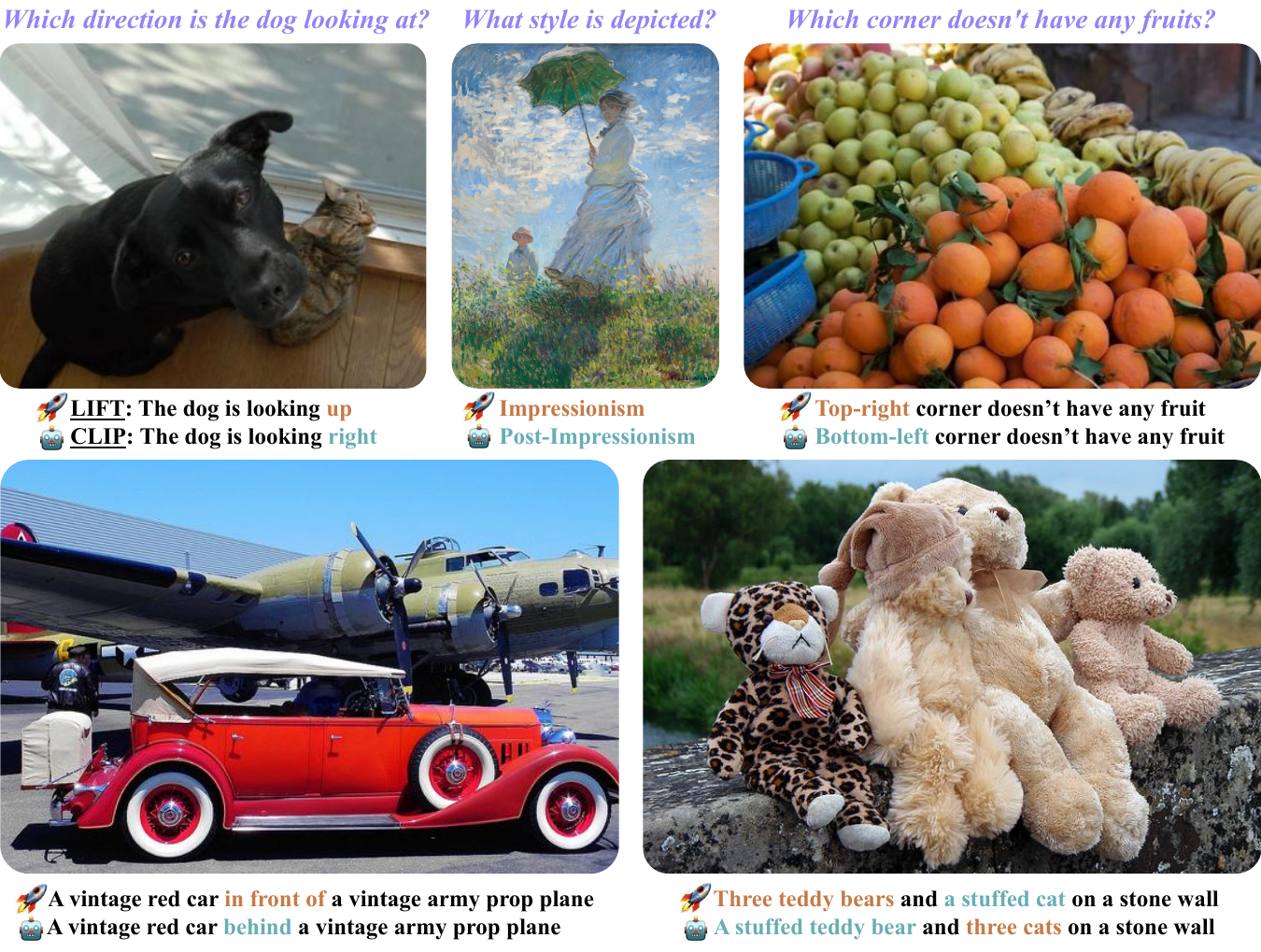}
      \caption{The qualitative comparisons between \ours and CLIP~\cite{radford2021learningtransferablevisualmodels}. The first line shows the caption or option selected by \ours, and the second line shows the one selected by CLIP.
      In every case, \ours selects the correct one, while CLIP does not. We observe that \ours compensates for CLIP’s shortcomings in tasks involving compositional information (e.g., spatial locations, object-attribute associations, object-object relations).}
      \label{fig:IntroDemo}
    \end{figure*}
}

\section{Introduction}
\label{sec:Introduction}
\vspace{-2mm}

\def\Pipeline#1{
    \captionsetup[sub]{font=small}
    \begin{figure*}[#1]
      \centering
      \includegraphics[width=0.99\linewidth]{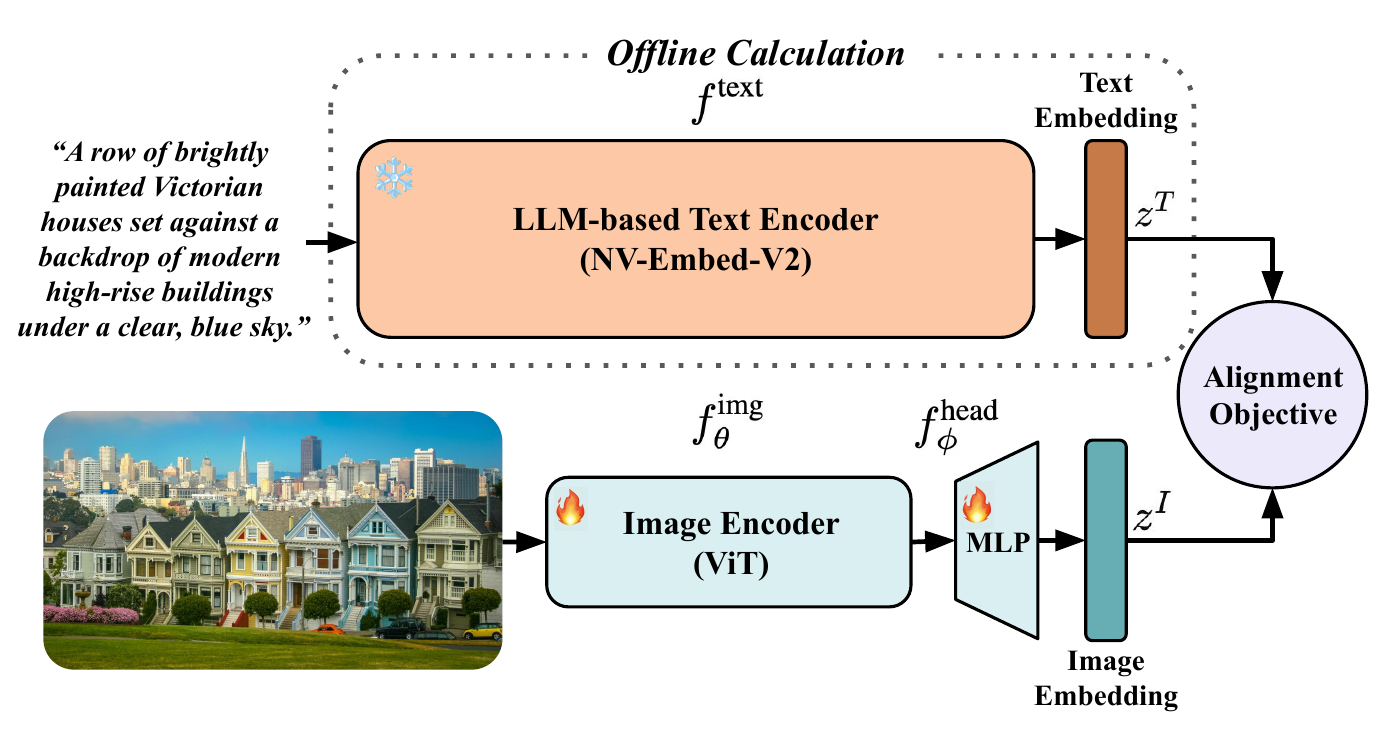}\vspace{-5pt}
      \caption{The pipeline of \ours, which adopts a dual-encoder architecture similar to CLIP~\cite{radford2021learningtransferablevisualmodels}. \ours uses an LLM-based text encoder $f^{\text{text}}$ to pre-compute the embedding $z^T$ for each text sample $T$ \textit{offline}. During training, we solely update the image encoder $f_{\theta}^{\text{img}}$ and the projection head $f_{\phi}^{\text{head}}$ to align image embeddings with the pre-computed text embeddings by optimizing an alignment objective.\vspace{-2mm}}
      \label{fig:pipeline}
    \end{figure*}
}

Contrastive pre-training on massive collections of text-image pairs has recently emerged as a powerful paradigm for learning language-aligned visual representations and demonstrates strong performance in applications such as vision-language models (VLMs)~\cite{liu2024improvedbaselinesvisualinstruction, liu2023visualinstructiontuning}. Representative works such as CLIP~\cite{radford2021learningtransferablevisualmodels} and SigLIP~\cite{zhai2023sigmoidlosslanguageimage} align text and image embeddings in a shared space by training separate text and image encoders on paired data with a contrastive loss. This approach is shown to be highly effective --- most VLMs today employ such pre-trained image encoders due to their language-aligned visual representations.

Despite its wide adoption, contrastive pre-training on text-image pairs has some widely known limitations. For instance, training both encoders from scratch makes the method computationally expensive~\cite{huang2024classificationrightvisionlanguagepretraining, mehta2024catlipcliplevelvisualrecognition}. This issue is further exacerbated by the large batch sizes and the substantial amounts of training data demanded by CLIP. Moreover, CLIP's text and image encoders struggle to accurately encode compositional information, including word order (in texts), spatial locations (in images), object–attribute associations (both), and object-object relations (both)~\cite{hsieh2023sugarcrepe, ma2023crepevisionlanguagefoundationmodels, thrush2022winogroundprobingvisionlanguage}. Prior studies attribute this limitation to the fact that contrastive pre-training on general-purpose retrieval datasets incentivizes CLIP's encoders trained from scratch to adopt a shortcut strategy that suppresses (i.e., discards) features related to compositional information~\cite{Geirhos_2020, robinson2021contrastivelearningavoidshortcut}. 

Crucially, a core assumption held by dominant contrastive approaches lies at the heart of these limitations: optimal language-image alignment requires jointly training text and image encoders from scratch. In this work, we question the necessity of this joint training and prove that large language models (LLMs) already provide good enough text embeddings to guide visual representation learning. Concretely, we use a pre-trained text encoder fine-tuned on an LLM to embed texts \textit{offline} and solely train the image encoder to align visual representations with the text embeddings. We name this approach \textbf{L}anguage-\textbf{I}mage alignment with a \textbf{F}ixed \textbf{T}ext encoder \textbf{(LIFT)}. The name ``LIFT'' also suggests that it aligns raw visual inputs with their corresponding higher-level semantics. \cref{fig:pipeline} illustrates the overall pipeline of the proposed framework.

To investigate whether, when, and why \ours might offer advantages over vanilla CLIP, we perform extensive benchmarking and ablation studies to address four fundamental questions:

\IntroDemo{t}

\begin{itemize}[leftmargin=14pt]
    \item \textit{Across tasks evaluating different model capabilities, on which does \ours demonstrate strengths compared to CLIP, and on which does it fall short?} \par
    \vspace{2pt}
    \textit{Answer}: \ours outperforms CLIP by an average accuracy gain of \textbf{7.4\%} across seven compositional understanding tasks and also leads CLIP on five out of six LLaVA~\cite{liu2024improvedbaselinesvisualinstruction, liu2023visualinstructiontuning}
    downstream tasks, all driven by its superior ability to encode compositional information. It also matches CLIP's zero-shot retrieval performance. Some qualitative results are shown in \cref{fig:IntroDemo}.
    \vspace{3pt}
    \item \textit{On which types of training data does \ours outperform CLIP, and why?} \par
    \vspace{2pt}
    \textit{Answer}: When trained on short, web-scraped captions, CLIP has a slight edge over \ours on three zero-shot retrieval tasks and one LLaVA downstream task. However, all of these advantages transfer to \ours when both are trained on long, synthetic captions. We attribute \ours's better performance to its robustness against the \textit{inverse effect} induced by synthetic captions.
    \vspace{3pt}
    \item \textit{What design choices of LLM-based text encoders enable better language-image alignment?} \par
    \vspace{2pt}
    \textit{\textit{Answer}}: Vanilla LLMs generally perform poorly as the text encoder for \ours. Contrastive fine-tuning targeted at improving text encoding is typically necessary, whereas additional embedding extraction modules are not.
    \vspace{3pt}
    \item \textit{Given its powerful LLM-based text encoder, can \ours simplify some of the design choices in mainstream contrastive language-image alignment approaches?} \par
    \vspace{2pt}
    \textit{Answer}: We find that a simpler yet efficient cosine similarity loss can substitute for the contrastive loss while achieving comparable performance on the compositional understanding tasks and LLaVA downstream tasks.
    
\end{itemize}
\Pipeline{t}

%% file: sections/2_related_works.tex
\def\flopsMemory#1{
    \begin{figure}[htbp]
    \centering
    \begin{minipage}[t]{0.48\textwidth}
        \centering
        \begin{tikzpicture}[scale=0.85]
            \definecolor{small_color}{HTML}{fcc8a5}
            \definecolor{base_color}{HTML}{d9eff2}
            \definecolor{large_color}{HTML}{ebe9fa}
            
            \begin{axis}[
                ybar,
                ylabel={FLOPs (G)},
                ylabel style={at={(axis description cs:0.2,1.02)}, anchor=south, rotate=270, font=\itshape\small},
                nodes near coords,
                nodes near coords align={vertical},
                bar width=10pt,
                xmin=0, xmax=1,
                ymin=0, ymax=580,
                xtick={0.21, 0.5, 0.79},
                xticklabels={CLIP(77), CLIP(128), \ours(77\text{,}128)},
                ytick={100, 200, 300, 400, 500, 600},
                legend style={at={(0.87,0.97)}, anchor=north, legend columns=1, draw=none, fill=none}
            ]
            \addplot [fill=small_color, draw=none] coordinates {(0.20,46) (0.49,59) (0.78,27)};
            \addplot [fill=base_color, draw=none] coordinates {(0.21,134) (0.50,155) (0.79,105)};
            \addplot [fill=large_color, draw=none] coordinates {(0.22,425) (0.51,464) (0.80,368)};
            \legend{ViT-S, ViT-B, ViT-L} 
            \end{axis}
        \end{tikzpicture}
        \caption{\small The estimated training FLOPs per text-image sample for CLIP~\cite{radford2021learningtransferablevisualmodels} and \ours trained with average per-batch max caption length 77 and 128.}
        \label{fig:flops}
    \end{minipage}
    \hfill
    \begin{minipage}[t]{0.48\textwidth}
        \centering
        \begin{tikzpicture}[scale=0.85]
            \definecolor{small_color}{HTML}{fcc8a5}
            \definecolor{base_color}{HTML}{d9eff2}
            \definecolor{large_color}{HTML}{ebe9fa}
            
            \begin{axis}[
                ybar,
                ylabel={Memory Usage (GB)},
                ylabel style={at={(axis description cs:0.3,1.02)}, anchor=south, rotate=270, font=\itshape\small},
                nodes near coords,
                nodes near coords align={vertical},
                bar width=10pt,
                xmin=0, xmax=1,
                ymin=0, ymax=57,
                xtick={0.21, 0.5, 0.79},
                xticklabels={CLIP(77), CLIP(128), \ours (77\text{,}128)},
                ytick={10, 20, 30, 40, 50},
                legend style={at={(0.87,0.97)}, anchor=north, legend columns=1, draw=none, fill=none}
            ]
            \addplot [fill=small_color, draw=none] coordinates {(0.20,14) (0.49,15) (0.78,13)};
            \addplot [fill=base_color, draw=none] coordinates {(0.21,20) (0.50,22) (0.79,19)};
            \addplot [fill=large_color, draw=none] coordinates {(0.22,36) (0.51,37) (0.80,33)};
            \legend{ViT-S, ViT-B, ViT-L} 
            \end{axis}
        \end{tikzpicture}
        \caption{\small The estimated H800 GPU memory usage for CLIP and \ours trained with average per-batch max caption length 77 and 128. The batch size is 1024.}
        \label{fig:memory}
    \end{minipage}
    \label{fig:comparison}
    \end{figure}
}

\section{Related Work}

\subsection{Language-Image Representation Learning}
Currently, language-image alignment is typically achieved through contrastive learning. Seminal works CLIP \cite{radford2021learningtransferablevisualmodels} and ALIGN \cite{jia2021scalingvisualvisionlanguagerepresentation} pioneer large-scale pre-training on text-image pairs to learn joint embeddings, enabling zero-shot transfer to various downstream tasks. These approaches rely on a dual-encoder architecture in which text and image encoders are jointly trained from scratch to maximize the alignment (e.g., measured by cosine similarity) between paired samples' embeddings while minimizing the alignment between non-matching pairs' embeddings. Subsequently, SigLIP \cite{zhai2023sigmoidlosslanguageimage} improves training stability using a sigmoid contrastive loss function. However, these methods demand significant computational resources to train both encoders.

Recently, SuperClass \cite{huang2024classificationrightvisionlanguagepretraining} and CatLIP \cite{mehta2024catlipcliplevelvisualrecognition} have explored an alternative training paradigm without text encoders. They extract class labels from captions and cast language-image alignment as a classification problem by training image encoders with a binary cross-entropy loss. However, since these approaches ignore word order in captions, the resulting representations behave like a bag-of-words and lack compositional understanding.

In this work, we aim to simplify the language-image alignment pipeline while avoiding the shortcomings of existing text encoder-free approaches.

\flopsMemory{t}

\subsection{The Limitations of CLIP}
It is well known that CLIP lacks compositional understanding, largely because contrastive pre-training on general-purpose retrieval datasets encourages CLIP's encoders to adopt a shortcut strategy that ignores compositional information~\cite{Geirhos_2020, yuksekgonul2023when}. To tackle this issue, \cite{yuksekgonul2023when} incorporates designed negative samples during training; \cite{tong2024cambrian1fullyopenvisioncentric, tong2024eyes} combine features from multiple image encoders. More recently, \cite{ maninis2025tipstextimagepretrainingspatial, mu2021slipselfsupervisionmeetslanguageimage, naeem2023silcimprovingvisionlanguage, tschannen2025siglip2multilingualvisionlanguage} merge self-supervised learning methods with contrastive learning, leading to image encoders with stronger vision-centric capabilities. Closer to our study, \cite{chen2024internvlscalingvisionfoundation, stone2025learningvisualcompositionimproved} replace CLIP’s text encoder with LLMs and jointly train them with image encoders, achieving better scaling behavior and improved compositional understanding. However, by introducing additional modifications to their training pipelines, they fail to isolate the use of unaltered LLM text embeddings and therefore do not systematically study their effect on language-image alignment.

Furthermore, \cite{liu2024clipsenhancedclipframework, liu2024mllmsaugmentedvisuallanguagerepresentationlearning, wang2023largedatareductionvisionlanguage} reveal that CLIP yields suboptimal zero-shot performance when trained on full-length long captions, which are typically synthesized by VLMs to include detailed object descriptions. \cite{liu2024mllmsaugmentedvisuallanguagerepresentationlearning} conjectures that CLIP's text encoder is distracted by the syntactic similarity of synthetic captions and fails to attend to semantically meaningful content. Truncation strategies, such as text shearing~\cite{liu2024mllmsaugmentedvisuallanguagerepresentationlearning} and sub-caption sampling~\cite{zheng2024dreamliplanguageimagepretraininglong}, can bring empirical improvements. However, they inevitably sacrifice the rich information that long captions provide.

In summary, current approaches remain insufficient in resolving both issues. We attempt to tackle both by questioning a prevailing setup in contrastive language-image alignment.

\subsection{Text Embedding Models}
Text embedding models are widely used to extract semantic embeddings of texts and are crucial in tasks such as Retrieval-Augmented Generation (RAG)~\cite{lewis2021retrievalaugmentedgenerationknowledgeintensivenlp}. Traditional embedding models such as BERT \cite{devlin2019bertpretrainingdeepbidirectional} and T5~\cite{raffel2023exploringlimitstransferlearning} are typically trained with bidirectional attention mechanisms. Recently, LLM-based embedding models such as LLM2Vec~\cite{behnamghader2024llm2veclargelanguagemodels}, E5-Mistral~\cite{wang2024improvingtextembeddingslarge}, SFR-Embedding-Mistral~\cite{SFRAIResearch2024}, and NV-Embed-V2~\cite{lee2025nvembedimprovedtechniquestraining} aim to leverage the rich semantics of auto-regressive LLMs and have emerged as powerful embedding models on Massive Text Embedding Benchmark (MTEB)~\cite{muennighoff2023mtebmassivetextembedding}. In particular, NV-Embed-V2~\cite{lee2025nvembedimprovedtechniquestraining} introduces a novel \textit{latent attention layer} and a two-stage contrastive instruction tuning, making the resulting text embeddings rich yet distinct between texts with different semantic meanings. Hence, it is adopted as the text encoder of \ours.

%% file: sections/3_method.tex
\def\NegCaptionDemo#1{
    \captionsetup[sub]{font=small}
    \begin{figure*}[#1]
      \centering
      \includegraphics[width=0.99\linewidth]{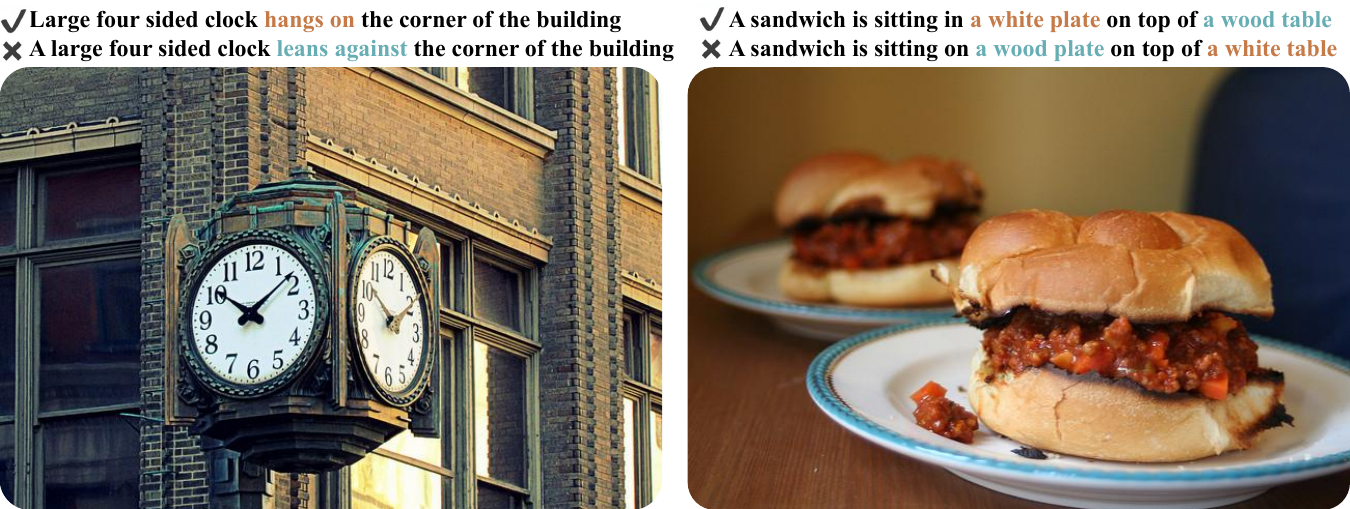}
      \caption{The original captions (top) and their negative counterparts (bottom) from two SugarCrepe~\cite{hsieh2023sugarcrepe} tasks: \texttt{replace relation} (left) and \texttt{swap attribute} (right).}
      \vspace{-8pt}
      \label{fig:NegCaptionDemo}
    \end{figure*}
}

\vspace{6pt}
\section{Method}

As illustrated in \cref{fig:pipeline}, \ours adopts a dual-encoder architecture similar to CLIP~\cite{radford2021learningtransferablevisualmodels}. Let \((T, I)\) denotes a text-image pair, where \(T \in \mathcal{V}^L\) is a sequence of \(L\) tokens from a vocabulary \(\mathcal{V}\) and \(I \in \mathbb{R}^{C \times H \times W}\). We apply a pre-trained and frozen LLM-based text encoder $f^{\text{text}}$ with an output dimension \(D\) to extract the text embedding 
\[z^T = f^{\text{text}}(T) \in \mathbb{R}^{1 \times D}.\]
\vspace{6pt}
We parameterize an image encoder neural network \( f_{\theta}^{\text{img}}: \mathbb{R}^{C \times H \times W} \rightarrow \mathbb{R}^{1 \times d} \) and a projection head \( f_{\phi}^{\text{head}}: \mathbb{R}^{1 \times d} \rightarrow \mathbb{R}^{1 \times D} \). \(f_{\theta}^{\text{img}}\) is implemented by a ViT~\cite{dosovitskiy2021imageworth16x16words} of width \(d\) and \( f_{\phi}^{\text{head}}\) is a 2-layer MLP. Let 
\[ z^I = f_{\phi}^{\text{head}} \circ f_{\theta}^{\text{img}}(I) \in \mathbb{R}^{1 \times D}\]
\vspace{6pt}be the final image embedding. We use CLIP's contrastive loss\footnote{In practice (see \cref{sec:4.4}), we find a simple \textit{cosine similarity loss} also leads to reasonably strong performance.}~\cite{radford2021learningtransferablevisualmodels, oord2019representationlearningcontrastivepredictive} as our objective, which is calculated between a batch of normalized \(z^T\) and \(z^I\) as:

\[\mathcal{L}_{\text{contrastive}} =-\frac{1}{2B} \sum_{i=1}^{B} \left[ \log \frac{\exp(z^T_i \cdot z^I_i / \tau)}{\sum_{j=1}^{B} \exp(z^T_i \cdot z^I_j / \tau)} + \log \frac{\exp(z^I_i \cdot z^T_i / \tau)}{\sum_{j=1}^{B} \exp(z^I_i \cdot z^T_j / \tau)} \right].\]

\NegCaptionDemo{t}

\vspace{12pt}
\textbf{Offline Text Embeddings Generation}. Since we don't optimize \(f^{\text{text}}\), the entire text embedding process can be performed \textit{offline}. Specifically, before training \ours, we embed all the captions in our dataset once, allowing subsequent trainings to reuse the pre-computed caption embeddings without the text encoder. \ours utilizes NV-Embed-V2~\cite{lee2025nvembedimprovedtechniquestraining} as \(f^{\text{text}}\). As a point of reference, eight H800 GPUs can embed 100M captions per day at \texttt{bfloat16} precision.

As shown in \cref{fig:flops} and \cref{fig:memory}, embedding texts offline leads to significant computational and memory efficiency. Concretely, given average per-batch max caption token length $n$, the FLOPs and memory footprint of CLIP scale with \(\mathcal{O}(n^2)\) complexity, whereas \ours achieves \(\mathcal{O}(1)\) amortized complexity. We also quantitatively benchmark CLIP and \ours on both short ($n=77$) and long ($n=128$) captions. On average, \ours reduces FLOPs by \textbf{25.5\%} for short captions and \textbf{35.7\%} for long ones, while lowering memory usage by \textbf{6.8\%} and \textbf{12.6\%}. The calculation details of FLOPs and memory usage are attached in \cref{sec:FLOPsCal}.

%% file: sections/4_experiements.tex
\def\Compositionality#1{
    \begin{table}[#1]
        \renewcommand{\arraystretch}{1.5} 
        \setlength{\arrayrulewidth}{2pt} 
        \setlength{\tabcolsep}{12pt}
        \centering
        \begin{adjustbox}{width=\textwidth}
        \begin{tabular}{lccccccccc}
        \Xhline{1.6pt}
        
        \multirow{2}{*}{Method} & \multirow{2}{*}{Dataset} & \multirow{2}{*}{\shortstack{Sample\\Seen}} & \multicolumn{2}{c}{Add} & \multicolumn{3}{c}{Replace} & \multicolumn{2}{c}{Swap}
        \vspace{-3pt} \\

        \cmidrule(lr{0.15em}){4-5}\cmidrule(lr{0.15em}){6-8}\cmidrule(lr{0.15em}){9-10}
        \specialrule{0pt}{-3pt}{0pt}
        & & & Obj & Att & Obj & Att & Rel & Obj & Att \\
        \Xhline{1.6pt}
        
        OpenCLIP & DataComp & 1.28B & 82.3 & 73.7 & 91.7 & 79.4 & 61.2 & 59.6 & 56.9 \\
        \rowcolor{gray!10}
        \ours & DataComp & 1.28B & \textbf{89.0} & \textbf{86.1} & \textbf{93.2} & \textbf{86.0} & \textbf{70.6} & \textbf{64.1} & \textbf{63.4} \\
        
        \Xhline{1.6pt}
        
        OpenCLIP & Recap & 512M & 77.0 & 73.7 & 88.9 & 80.8 & 63.4 & 62.0 & \textbf{76.3} \\
        \rowcolor{gray!10}
        \ours & Recap & 512M & \textbf{88.8} & \textbf{92.2} & \textbf{92.3} & \textbf{88.2} & \textbf{76.8} & \textbf{66.5} & 72.8 \\
        
        \Xhline{1.6pt}
        \end{tabular}
        \end{adjustbox}
        \vspace{6pt}
        \caption{\small The performance of \ours and CLIP~\cite{radford2021learningtransferablevisualmodels} on seven SugarCrepe~\cite{hsieh2023sugarcrepe} tasks. For each task, SugarCrepe generates negative captions by \texttt{add}, \texttt{replace}, or \texttt{swap} an \texttt{object}, \texttt{attribute}, or \texttt{relation} in the original captions. We report the accuracy of each task, and the best results are bolded.}
        \vspace{-20pt}
        \label{tab:Compositionality}
    \end{table}
}

\def\MLLM#1{
    \begin{table}[#1]
        \renewcommand{\arraystretch}{1.5} 
        \setlength{\arrayrulewidth}{2pt} 
        \setlength{\tabcolsep}{11pt}
        \centering
        \begin{adjustbox}{width=\textwidth}
        \begin{tabular}{lcccccccc}
        \Xhline{1.6pt}
        
        \multirow{2}{*}{Method} & \multirow{2}{*}{Dataset} & \multirow{2}{*}{\shortstack{Sample\\Seen}} & \multirow{2}{*}{TextVQA} & \multicolumn{2}{c}{MMBench} & \multirow{2}{*}{MME} & \multirow{2}{*}{POPE} & \multirow{2}{*}{SciQA}
        \vspace{-3pt} \\

        \cmidrule(lr{0.15em}){5-6}
        \specialrule{0pt}{-3pt}{0pt}
        & & & & EN & CN & & & \\
        \Xhline{1.6pt}
        
        OpenCLIP & DataComp & 1.28B & 47.9 & 53.3 & 45.9 & 1283.2 & \textbf{86.4} & 69.9 \\
        \rowcolor{gray!10}
        \ours & DataComp & 1.28B & \textbf{48.9} & \textbf{57.8} & \textbf{50.6} & \textbf{1289.2} & 85.2 & \textbf{70.2} \\
        \Xhline{1.6pt}

        OpenCLIP & Recap & 512M & 46.2 & 48.4 & 42.0 & 1245.1 & 85.6 & 69.0 \\
        \rowcolor{gray!10}
        \ours & Recap & 512M & \textbf{47.8} & \textbf{54.3} & \textbf{46.5} & \textbf{1341.6} & \textbf{85.8} & \textbf{69.4} \\
        \Xhline{1.6pt}
        
        \end{tabular}
        \end{adjustbox}
        \vspace{6pt}
        \caption{\small The performance of the LMM with either \ours or CLIP~\cite{radford2021learningtransferablevisualmodels} as the vision tower on LLaVA~\cite{liu2024improvedbaselinesvisualinstruction, liu2023visualinstructiontuning} downstream tasks. The results are reported for TextVQA~\cite{singh2019vqamodelsread}, MMBench~\cite{liu2024mmbenchmultimodalmodelallaround} (English and Chinese), MME~\cite{fu2024mmecomprehensiveevaluationbenchmark}, POPE~\cite{li2023evaluatingobjecthallucinationlarge} (random accuracy), and ScienceQA~\cite{lu2022learnexplainmultimodalreasoning} (accuracy). The best results are bolded.}
        \vspace{-10pt}
        \label{tab:MLLM}
    \end{table}
}

\def\MMBench#1{
    \begin{table}[#1]
        \renewcommand{\arraystretch}{1.5} 
        \setlength{\arrayrulewidth}{2pt} 
        \setlength{\tabcolsep}{13pt}
        \centering
        \begin{adjustbox}{width=\textwidth}
        \begin{tabular}{lcccccccc}
        \Xhline{1.6pt}
        
        \multirow{2}{*}{Method} & \multirow{2}{*}{Dataset} & \multirow{2}{*}{\shortstack{Sample\\Seen}} & \multirow{2}{*}{AR} & \multirow{2}{*}{CP} & \multirow{2}{*}{FP-C} & \multirow{2}{*}{FP-S} & \multirow{2}{*}{LR} & \multirow{2}{*}{RR}
        \vspace{-3pt} \\

        & & & & & & & & \\
        \Xhline{1.6pt}
        
        OpenCLIP & DataComp & 1.28B & 61.3 & 65.9 & 50.3 & 52.6 & 27.1 & 39.1 \\
        \rowcolor{gray!10}
        \ours & DataComp & 1.28B & \textbf{67.8} & \textbf{72.3} & \textbf{51.0} & \textbf{56.0} & \textbf{28.0} & \textbf{47.0} \\
        \Xhline{1.6pt}

        OpenCLIP & Recap & 512M & 58.3 & 63.5 & 45.5 & 43.0 & 26.3 & 32.2 \\
        \rowcolor{gray!10}
        \ours & Recap & 512M & \textbf{60.0} & \textbf{70.6} & \textbf{50.3} & \textbf{51.9} & \textbf{28.0} & \textbf{40.9} \\
        \Xhline{1.6pt}
        
        \end{tabular}
        \end{adjustbox}
        \vspace{6pt}
        \caption{\small The performance of the LMM with either \ours or CLIP~\cite{radford2021learningtransferablevisualmodels} as the vision tower on specific MMBench~\cite{liu2024mmbenchmultimodalmodelallaround} subtasks. Abbreviations: AR for \texttt{Attribute Reasoning}; CP for \texttt{Coarse Perception}; FP-C for
        \texttt{Fine-grained Perception (Cross Instance)}; FP-S for \texttt{Fine-grained Perception (Single Instance)}; LR for \texttt{Logical Reasoning}; RR for \texttt{Relation Reasoning}. The visualizations are provided in \cref{sec:MMBenchTasksVisualizations}. The best results are bolded.}
        \vspace{-10pt}
        \label{tab:MMBench}
    \end{table}
}

\def\Zeroshot#1{
    \begin{table}[h]
        \renewcommand{\arraystretch}{1.5} 
        \setlength{\arrayrulewidth}{2pt} 
        \setlength{\tabcolsep}{14pt}       
        \centering
        \resizebox{\textwidth}{!}{ 
        \begin{tabular}{lcc>{\centering\arraybackslash}p{1.5cm}cccc}
        \Xhline{1.6pt}
        
        \multirow{2}{*}{Method} & \multirow{2}{*}{Dataset} & \multirow{2}{*}{\shortstack{Sample\\Seen}} & \multirow{2}{*}{ImageNet} & \multicolumn{2}{c}{COCO} & \multicolumn{2}{c}{Flickr}
        \vspace{-3pt} \\

        \cmidrule(lr{0.15em}){5-6} \cmidrule(lr{0.15em}){7-8} 
        \specialrule{0pt}{-3pt}{0pt}
        & & & & I2T & T2I & I2T & T2I \\
        \Xhline{1.6pt}

        OpenCLIP & DataComp & 1.28B & \textbf{58.4} & \textbf{31.0} & 27.2 & \textbf{62.9} & 59.6 \\
        \rowcolor{gray!10}
        \ours & DataComp & 1.28B & 58.3 & 29.1 & \textbf{28.1} & 58.8 & \textbf{63.7} \\
        \Xhline{1.6pt}

        OpenCLIP & Recap & 512M & 34.6 & 25.7 & 26.7 & 56.4 & 57.9 \\
        \rowcolor{gray!10}
        \ours & Recap & 512M & \textbf{43.6} & \textbf{34.6} & \textbf{36.0} & \textbf{69.1} & \textbf{72.9} \\
        
        \Xhline{1.6pt}
        \end{tabular}
        }
        \vspace{6pt}
        \caption{\small The zero-shot performance of \ours and CLIP~\cite{radford2021learningtransferablevisualmodels} on ImageNet-1K~\cite{5206848} classification, COCO~\cite{lin2015microsoftcococommonobjects} Image-to-Text and Text-to-Image retrieval, and Flickr30K~\cite{young-etal-2014-image} Image-to-Text and Text-to-Image retrieval. For all tasks, we report top-1 accuracy. The best results are bolded.}
        \vspace{-8pt}
        \label{tab:Zeroshot}
    \end{table}
}

\def\HomoCaptionDemo#1{
    \captionsetup[sub]{font=small}
    \begin{figure*}[#1]
      \centering
      \includegraphics[width=0.99\linewidth]{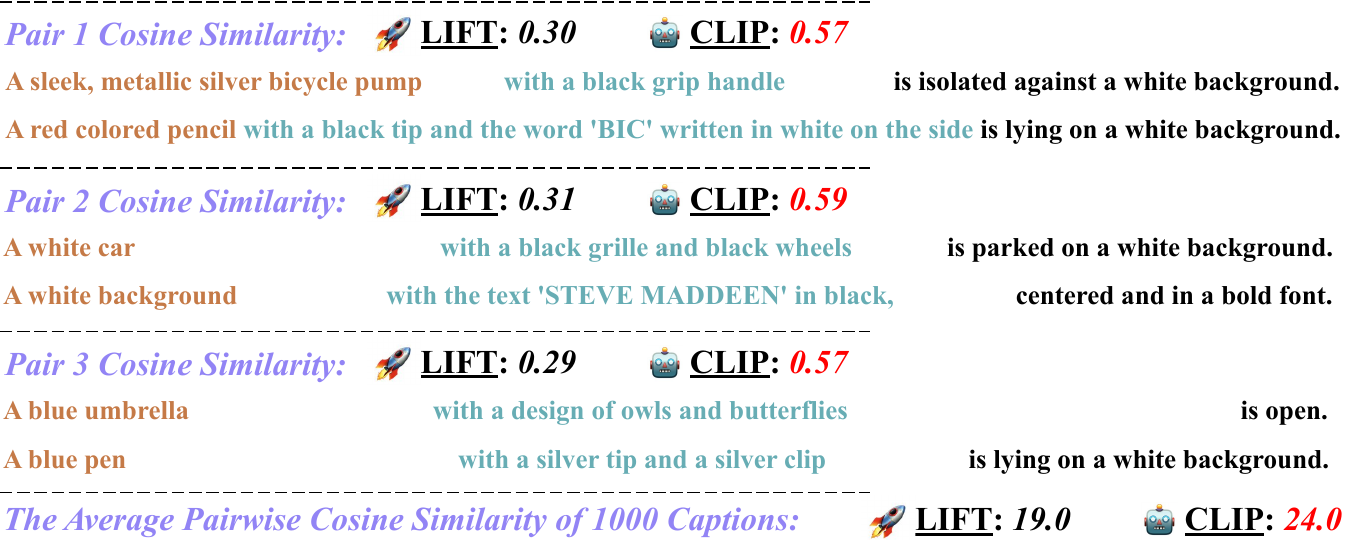}
      \caption{The examples of syntactically similar but semantically different caption pairs from Recap-DataComp-1B~\cite{li2024recaptionbillionswebimages}. The synthetic captions follow the template ``\texttt{A \{Adj.\} \{Noun\} with \{N.~Phrase\} \{Verb\} \{Location\}}''. CLIP's~\cite{radford2021learningtransferablevisualmodels} text encoder often assigns higher scores to caption pairs with similar syntax, while \ours better captures semantic differences and assigns lower ones. The scores are calculated based on the embeddings from LIFT's $f^{\text{text}}$ and CLIP's text encoder trained on 512M Recap-DataComp-1B samples.}
      \vspace{-15pt}
      \label{fig:HomoCaptionDemo}
    \end{figure*}
}

\def\EmbeddingModel#1{
    \begin{table}[#1]
        \renewcommand{\arraystretch}{1.5} 
        \setlength{\arrayrulewidth}{1pt} 
        \setlength{\tabcolsep}{11pt}       
        \centering
        \resizebox{\textwidth}{!}{ 
        \begin{tabular}{lcccccccccc}
        \Xhline{1.6pt}
        
        \multirow{2}{*}{LLM Encoder} & \multirow{2}{*}{\shortstack{Contrastive\\Fine-tuning}} & \multirow{2}{*}{\shortstack{Embedding\\Extraction}} & \multirow{2}{*}{ImageNet} & \multicolumn{2}{c} {Flickr} & \multicolumn{3}{c} {Replace} \vspace{-3pt} \\

        \cmidrule(lr{0.15em}){5-6} \cmidrule(lr{0.15em}){7-9}
        \specialrule{0pt}{-3pt}{0pt}
        & & & & I2T & T2I & Obj & Att & Rel \\
        
        \Xhline{1.6pt}

        Mistral-7B-V0.1 & \xmark & \texttt{<eos>} & 12.5 & 5.4 & 3.3 & 69.0 & 58.4 & 50.0 \\ 
        Vicuna-7B-V1.5 & \xmark & \texttt{<eos>} & 23.8 & 14.1 & 13.2 & 78.1 & 73.1 & 58.8 \\
        \Xhline{1.6pt}
        
        SFR-Embed-Mistral & \cmark & \texttt{<eos>} & \underline{39.9} & \textbf{44.9} & \underline{42.1} & \textbf{87.4} & \textbf{82.3} & \textbf{70.9} \\
        Linq-Embed-Mistral & \cmark & \texttt{<eos>} & 39.8 & \underline{44.2} & 41.6 & 
        84.8 & \underline{81.5} & 64.2 \\
        NV-Embed-V2 & \cmark & \texttt{\shortstack{Latent Attention}} & \textbf{40.9} & 42.7 & \textbf{44.1} & \underline{87.2} & 81.2 & \underline{67.3} \\

        \Xhline{1.6pt}
        
        \end{tabular}
        }
        \vspace{4pt}
        \caption{\small The ablation study on the choice of \ours's LLM-based text encoder. The results are reported for ImageNet-1K~\cite{5206848} classification, Flickr30K~\cite{young-etal-2014-image} retrieval tasks, and SugarCrepe's~\cite{hsieh2023sugarcrepe} \texttt{replace} task. All models are trained on 128M samples from DataComp-1B~\cite{gadre2023DataCompsearchgenerationmultimodal}. The best results are bolded, and the second-best results are underlined.}
        \vspace{-10pt}
        \label{tab:EmbeddingModel}
    \end{table}
}

\def\CosVsContrastive#1{
    \begin{table}[#1]
        \renewcommand{\arraystretch}{1.5} 
        \setlength{\arrayrulewidth}{1pt} 
        \setlength{\tabcolsep}{10pt}       
        \centering
        \resizebox{\textwidth}{!}{ 
        \begin{tabular}{lcccccccccc}
        \Xhline{1.6pt}
        
        \multirow{2}{*}{Loss} & \multirow{2}{*}{Dataset} & \multirow{2}{*}{\shortstack{Sample\\Seen}} & \multirow{2}{*}{ImageNet} & \multicolumn{2}{c}{Flickr}
        & \multicolumn{2}{c}{Add} & \multicolumn{2}{c}{MMBench} 
        \vspace{-3pt} \\
        
        \cmidrule(lr{0.15em}){5-6} \cmidrule(lr{0.15em}){7-8} \cmidrule(lr{0.15em}){9-10}
        \specialrule{0pt}{-3pt}{0pt}
        & & & & I2T & T2I & Obj & Att & EN & CN \\
        
        \Xhline{1.6pt}
        
        Contrastive & DataComp & 426M & \textbf{45.5} & \textbf{45.8} & \textbf{51.9} & 82.4 & \textbf{84.4} & \textbf{54.0} & \textbf{43.4} \\
        Cosine Sim & DataComp & 426M & 26.8 & 10.2 & 19.5 & \textbf{85.4} & 74.0 & 53.4 & 43.0 \\

        \Xhline{1.6pt}

        Contrastive & Recap & 512M & \textbf{43.6} & \textbf{69.1} & \textbf{72.9} & \textbf{88.8} & \textbf{92.2} & 54.3 & 46.5 \\
        Cosine Sim & Recap & 512M & 38.8 & 58.8 & 68.9 & 86.9 & 88.2 & \textbf{57.5} & \textbf{50.3} \\

        \Xhline{1.6pt}
        
        \end{tabular}
        }
        \vspace{4pt}
        \caption{\small The ablation study on the choice of \ours's loss function. The results are reported for ImageNet-1K~\cite{5206848} classification, Flickr30K~\cite{young-etal-2014-image} retrieval tasks, SugarCrepe's~\cite{hsieh2023sugarcrepe} \texttt{Add} task, and MMBench~\cite{liu2024mmbenchmultimodalmodelallaround} (English and Chinese). The best results are bolded. }
        \vspace{-10pt}
        \label{tab:CosVsContrastive}
    \end{table}
}

\vspace{6pt}
\section{Experiments}
\label{sec:Experiments}
\textbf{Experimental Settings}. We train both \ours and CLIP~\cite{radford2021learningtransferablevisualmodels} (implemented by OpenCLIP~\cite{Cherti_2023}) using a ViT-B/16~\cite{dosovitskiy2021imageworth16x16words} vision backbone on a dataset containing 400 million text-image pairs. Each image has two types of captions: a short, web-scraped caption from DataComp-1B~\cite{gadre2023DataCompsearchgenerationmultimodal}, and a long, synthetic caption from Recap-DataComp-1B~\cite{li2024recaptionbillionswebimages}. Their respective max caption token lengths are set to 77 and 323. To ensure a fair comparison, both \ours and CLIP are trained using the same hyperparameters: 500 warmup steps, a weight decay of 0.2, a learning rate of 1e-3 with a cosine schedule, and a batch size of 16,384. We use an input resolution of 224 $\times$ 224. The largest experiment uses 1.28 billion samples and is trained on eight H800 GPUs over 12 days.

In the following sections, we systematically address the four questions raised in \cref{sec:Introduction}.
\vspace{-5pt}

\subsection{\textit{On which tasks does \ours offer advantages, and on which does it fall short?}}
\label{sec:4.1}

\textbf{Compositional Understanding}. Compositional understanding is a known limitation of CLIP. We evaluate it using seven SugarCrepe~\cite{hsieh2023sugarcrepe} tasks. As shown in \cref{fig:NegCaptionDemo}, for each caption, SugarCrepe generates a negative caption by \texttt{add}, \texttt{replace}, or \texttt{swap} an \texttt{object}, \texttt{attribute}, or \texttt{relation} in the original caption. Models are asked to identify the correct caption based on caption-image cosine similarity.
\Compositionality{h}

As shown in \cref{tab:Compositionality}, when trained on the short captions from DataComp-1B, \ours outperforms CLIP on all seven tasks with a \textbf{6.8\%} average accuracy gain; when trained on the long, synthetic captions from Recap-DataComp-1B, it leads on six tasks with a \textbf{7.9\%} gain. In both settings, \ours achieves significant gains on \texttt{add attribute}, \texttt{replace attribute}, and \texttt{replace relation} tasks. These improvements are strong evidence that $f^{\text{text}}$'s auto-regressive training objective avoids the compositional oversight induced by contrastive learning and enables more accurate modeling of object–attribute associations and object–object relations. More visualizations can be found in \cref{sec:SCMoreVisualizations}.

Admittedly, \ours’s ability to capture compositional information is not yet complete. It shows relatively low accuracy on \texttt{swap object} and \texttt{swap attribute} compared to other SugarCrepe tasks. We further attribute this limitation to the fact that the contrastive learning objective still focuses on aligning primarily lower-order statistics. Addressing this challenge requires exploring more refined information-theoretic measures for language-image alignment, a key direction for future work.

\textbf{LLaVA~\cite{liu2024improvedbaselinesvisualinstruction, liu2023visualinstructiontuning} Downstream Tasks}. \cite{tong2024eyes} shows that large multimodal models (LMMs) using CLIP as the vision tower inherit CLIP’s limitations in modeling certain visual patterns. We train LMMs using LLaVA to examine whether \ours transfers its superior compositional understanding to the LMM built on it. Vicuna-7B-V1.5~\cite{vicuna2023} is used as the base language model and is fine-tuned with LoRA~\cite{hu2021loralowrankadaptationlarge}. All training hyperparameters follow the original LLaVA setup, except that we initialize LLaVA's projector with the weights of our 2-layer MLP \(f^{\text{head}}_{\phi}\).

\MLLM{H}
\MMBench{H}

As shown in \cref{tab:MLLM}, \ours outperforms CLIP on five out of six LLaVA downstream tasks when both are trained on short captions, and leads on all tasks when trained on long captions. In both settings, \ours shows substantial improvements on MMBench~\cite{liu2024mmbenchmultimodalmodelallaround}. We further examine their performance on specific MMBench (English) subtasks to identify the exact visual patterns that offer the gains. As shown in \cref{tab:MMBench}, \ours achieves significant accuracy gains on \texttt{fine-grained perception (single-instance)} and \texttt{relational reasoning}. The former subtask involves object localization and attribute recognition, while the latter includes identifying physical relations, all largely benefiting from \ours's accurate encoding of compositional information. The visualizations are provided in \cref{sec:MMBenchTasksVisualizations}.

\textbf{Zero-shot Classification and Retrieval}. Similar to CLIP, \ours can perform zero-shot transfer for image classification and cross-modal retrieval by employing the LLM-based text encoder \(f^{\text{text}}\) to embed captions during inference. Following CLIP's evaluation protocol, we evaluate the models on ImageNet-1K validation set~\cite{5206848} by constructing image captions using the prompt template ``\texttt{It is a photo of \{label\}}.'' More evaluation details are provided in \cref{sec: DatasetInfo}.

As shown in \cref{tab:Zeroshot}, when trained on short captions, \ours outperforms CLIP on two text-to-image tasks, while performing similarly on the remaining three tasks. When trained on long captions, however, \ours leads CLIP by substantial margins on all these tasks, achieving an average accuracy gain of \textbf{11.0\%}. We analyze the cause of this decline in CLIP's relative performance in \cref{sec:4.2}.
\Zeroshot{H}

\vspace{-8pt}
\subsection{\textit{On which types of training data does \ours outperform CLIP, and why?}}
\label{sec:4.2}
\HomoCaptionDemo{t}
As reported in \cref{sec:4.1}, when trained on short captions, CLIP~\cite{radford2021learningtransferablevisualmodels} has a slight edge over \ours on POPE~\cite{li2023evaluatingobjecthallucinationlarge}, ImageNet-1K~\cite{5206848} zero-shot classification, and two image-to-text retrieval tasks. However, all of these advantages are overtaken by \ours when both are trained on long, synthetic captions. In this section, we investigate the reasons behind CLIP's loss of performance advantages.

One contributing factor is the \textit{inverse effect}~\cite{li2023inversescalinglawclip, liu2024clipsenhancedclipframework}, which observes that CLIP trained on full-length synthetic captions yields suboptimal zero-shot performance but shows noticeable improvements as the captions are progressively truncated. This effect likely stems from the homogeneous caption syntax introduced by caption generators (usually fine-tuned VLMs), which can distort the original caption distribution and become a ``shortcut'' feature~\cite{liu2024mllmsaugmentedvisuallanguagerepresentationlearning, robinson2021contrastivelearningavoidshortcut}. Such homogeneous syntax is most pronounced in full-length synthetic captions and weakens with increasing truncation.

As shown in \cref{fig:HomoCaptionDemo}, CLIP's text encoder is misled by this shortcut feature during training from scratch. By computing the average pairwise cosine similarity of 1,000 captions randomly drawn from Recap-DataComp-1B~\cite{li2024recaptionbillionswebimages}, we find that CLIP’s text encoder overemphasizes syntactic similarity, assigning high similarity scores to caption pairs that are syntactically similar but semantically different. In contrast, \ours employs the LLM-based encoder $f^{\text{text}}$ pre-trained on large-scale data, which yields an embedding space more robust to syntactic homogeneity. It focuses more on semantic content and assigns significantly lower similarity scores to such misleading caption pairs.

However, the 11.0\% accuracy gap between \ours and CLIP cannot be fully attributed to the inverse effect. We argue that another key factor is the expressive power of text encoders, which is especially critical for handling the greater semantic complexity of long, synthetic captions. Specifically, the CLIP's text encoder in our implementation has 63M parameters, whereas \ours's $f^{\text{text}}$ has 7B parameters. Despite its larger architecture, \ours remains more efficient for long-caption processing due to its use of offline embeddings.

\subsection{\textit{What design choices of LLM-based text encoders enable better language-image alignment?}}
Several key design choices have contributed to the transformation of LLMs into state-of-the-art text encoders, and this section examines which of these choices are most helpful for the language-image alignment achieved by \ours. \cite{morris2023languagemodelinversion, neelakantan2022textcodeembeddingscontrastive} first show that the hidden state of end-of-sentence token \texttt{<eos>} contains sufficient information about an input sequence and can be used as its embedding. Later methods explore alternative embedding extraction strategies, as \texttt{<eos>} tends to bias attention toward nearby tokens and ignore distant ones~\cite{jaegle2021perceivergeneralperceptioniterative, lee2025nvembedimprovedtechniquestraining}. Meanwhile, contrastive fine-tuning using instructions designed for general text embedding tasks (retrieval, clustering, etc.) has been shown to further improve embedding quality~\cite{SFRAIResearch2024, LinqAIResearch2024, lee2025nvembedimprovedtechniquestraining}.

To disentangle the effects of these design choices on language-image alignment, we train \ours using five representative LLMs as $f^{\text{text}}$. Mistral-7B-V0.1~\cite{jiang2023mistral7b} and Vicuna-7B-V0.1~\cite{vicuna2023} are vanilla LLMs without any fine-tuning or architectural changes; SFR-Embed-Mistral~\cite{SFRAIResearch2024} and Linq-Embed-Mistral~\cite{LinqAIResearch2024} both apply contrastive fine-tuning; NV-Embed-V2~\cite{lee2025nvembedimprovedtechniquestraining} represents the most evolved variant, combining contrastive fine-tuning with a latent attention layer to extract embeddings. All models have 7B parameters, with a hidden size and final embedding dimension of 4096.

As shown in \cref{tab:EmbeddingModel}, two vanilla LLMs lag significantly behind their fine-tuned counterparts across all reported tasks --- for instance, by an average accuracy decline of 22.8\% on ImageNet-1K~\cite{5206848} zero-shot classification. Vanilla Mistral-7B-V0.1 even performs no better than random guessing on SugarCrepe's~\cite{hsieh2023sugarcrepe} \texttt{replace relation} task. These results indicate that LLMs are not inherently effective as the text encoder for \ours, and contrastive fine-tuning is necessary. On the other hand, all three fine-tuned models achieve comparable performance, suggesting that \texttt{<eos>} token alone can accurately encode input captions, and that advanced embedding extraction mechanisms, such as NV-Embed-V2’s additional latent attention layer, may not be required.
\EmbeddingModel{H}

\subsection{\textit{Can \ours simplify some of the design choices in mainstream contrastive language-image alignment approaches?}}
\label{sec:4.4}
\textbf{A Simple Cosine Similarity Loss}. CLIP~\cite{radford2021learningtransferablevisualmodels} and its variants have demonstrated the effectiveness of cosine similarity for language-image alignment. To avoid mode collapse (i.e., identical outputs from text and image encoders regardless of the inputs), CLIP employs a contrastive InfoNCE~\cite{oord2019representationlearningcontrastivepredictive} loss based on pairwise cosine similarities. However, this approach is computationally intensive, with both FLOPs and memory scaling asymptotically as \(\mathcal{O}(B^2)\)  with batch size $B$. It also requires a large batch size to ensure sufficient negative samples. SigLIP~\cite{zhai2023sigmoidlosslanguageimage} introduces a chunked strategy that relaxes the large‑batch requirement, yet its FLOPs and memory still grow quadratically with local batch size.

Since the embedding space of \(f^{\text{text}}\) is fixed, mode collapse is no longer a concern. This motivates us to explore whether a simple cosine similarity loss, computed solely on positive text-image pairs without involving negative pairs, can be effective as well. Specifically, for a batch of size \(B\), we optimize
\[
\mathcal{L}_{\text{cosine}} = \frac{1}{B}\sum_{i=1}^{B}(1 - z^T_i \cdot z^I_i),
\]
where \(z^T = f^{\text{text}}(T)\) and \( z^I = f_{\phi}^{\text{head}} \circ f_{\theta}^{\text{img}}(I)\) is a normalized embedded text-image pair in the batch. This simple loss has \(\mathcal{O}(B)\) FLOPs and memory complexity with respect to batch size \(B\) and removes the reliance on negative samples, thereby easing the batch size constraint. \cite{chen2020exploringsimplesiameserepresentation, grill2020bootstraplatentnewapproach} also explore conceptually similar cosine similarity losses, but they employ more sophisticated techniques to address mode collapse.
\CosVsContrastive{H}

As shown in \cref{tab:CosVsContrastive}, the simple cosine similarity loss performs comparably to the contrastive loss on the compositional understanding tasks and LLaVA~\cite{liu2024improvedbaselinesvisualinstruction, liu2023visualinstructiontuning} downstream tasks. Notably, when trained on long captions, \ours using the simple cosine similarity loss outperforms its contrastive loss variant on both English and Chinese MMBench~\cite{liu2024mmbenchmultimodalmodelallaround} by non-trivial margins. However, it suffers significant performance drops in the zero-shot retrieval tasks, particularly when trained on short, web-scraped captions. The simple cosine similarity loss shows an accuracy decline of 18.7\% on ImageNet-1K~\cite{5206848} and 34.0\% on Flickr30K~\cite{young-etal-2014-image}. We attribute this performance gap to the contrastive loss’s use of negative samples, which encourages more discriminative representations that benefit classification and retrieval tasks.

%% file: sections/5_conclusion.tex
\section{Conclusion}
\label{sec:Conclusion}

In this paper, we question a core assumption held by dominant language-image alignment approaches like CLIP~\cite{radford2021learningtransferablevisualmodels} --- that text and image encoders should be jointly trained from scratch to achieve optimal language-image alignment. We present \ours, which pre-computes fixed text embeddings from LLMs and solely trains the image encoder. Comprehensive benchmarking and ablation studies confirm that \ours outperforms CLIP in key scenarios that involve compositional understanding and long captions. Further experiments show that contrastive fine-tuning is crucial for \ours's LLM-based text encoder to generate effective text embeddings and that \ours also enables the use of a simpler yet effective loss function. Our work initiates a systematic exploration of how text embeddings from LLMs can guide visual representation learning and opens a new avenue for language-image alignment.


%% file: sections/6_appendix.tex
\setcounter{section}{0}
\renewcommand{\thesection}{A.\arabic{section}}
\setcounter{equation}{0}
\renewcommand{\theequation}{A\arabic{equation}}
\setcounter{table}{0}
\renewcommand{\thetable}{A\arabic{table}}
\setcounter{figure}{0}
\renewcommand{\thefigure}{A\arabic{figure}}

\def\ScAppendixOne#1{
    \captionsetup[sub]{font=small}
    \begin{figure*}[#1]
      \centering
      \includegraphics[width=0.99\linewidth]{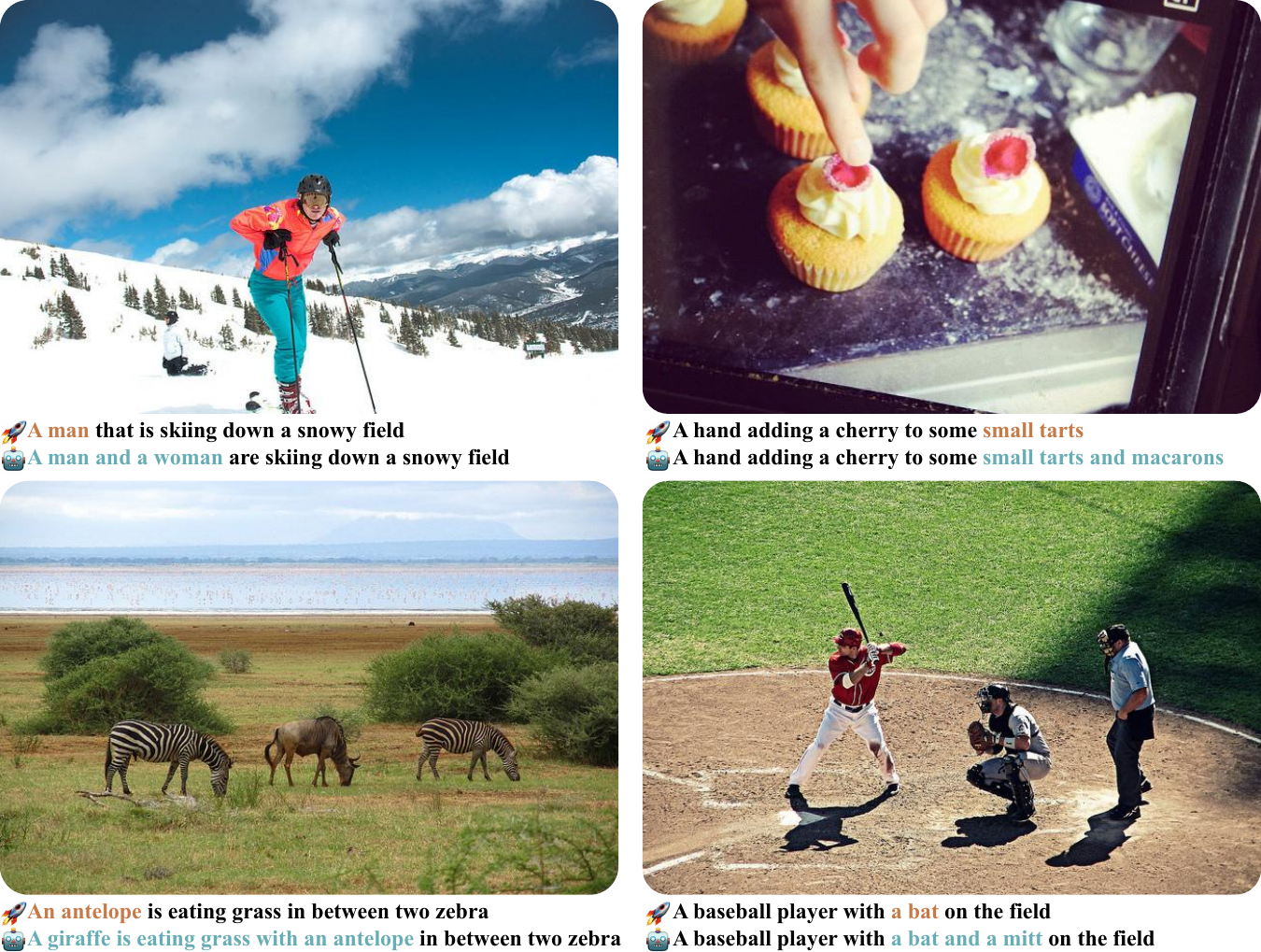}
      \caption{The visualizations of the \texttt{add object} task from SugarCrepe~\cite{hsieh2023sugarcrepe}. Each subfigure shows the captions selected by \ours (top) and CLIP~\cite{radford2021learningtransferablevisualmodels} (bottom). In every case, \ours selects the correct caption, while CLIP does not.}
      \vspace{-8pt}
      \label{fig:ScAppendixOne}
    \end{figure*}
}

\def\MMBAppendixOne#1{
    \captionsetup[sub]{font=small}
    \begin{figure*}[#1]
      \centering
      \includegraphics[width=0.99\linewidth]{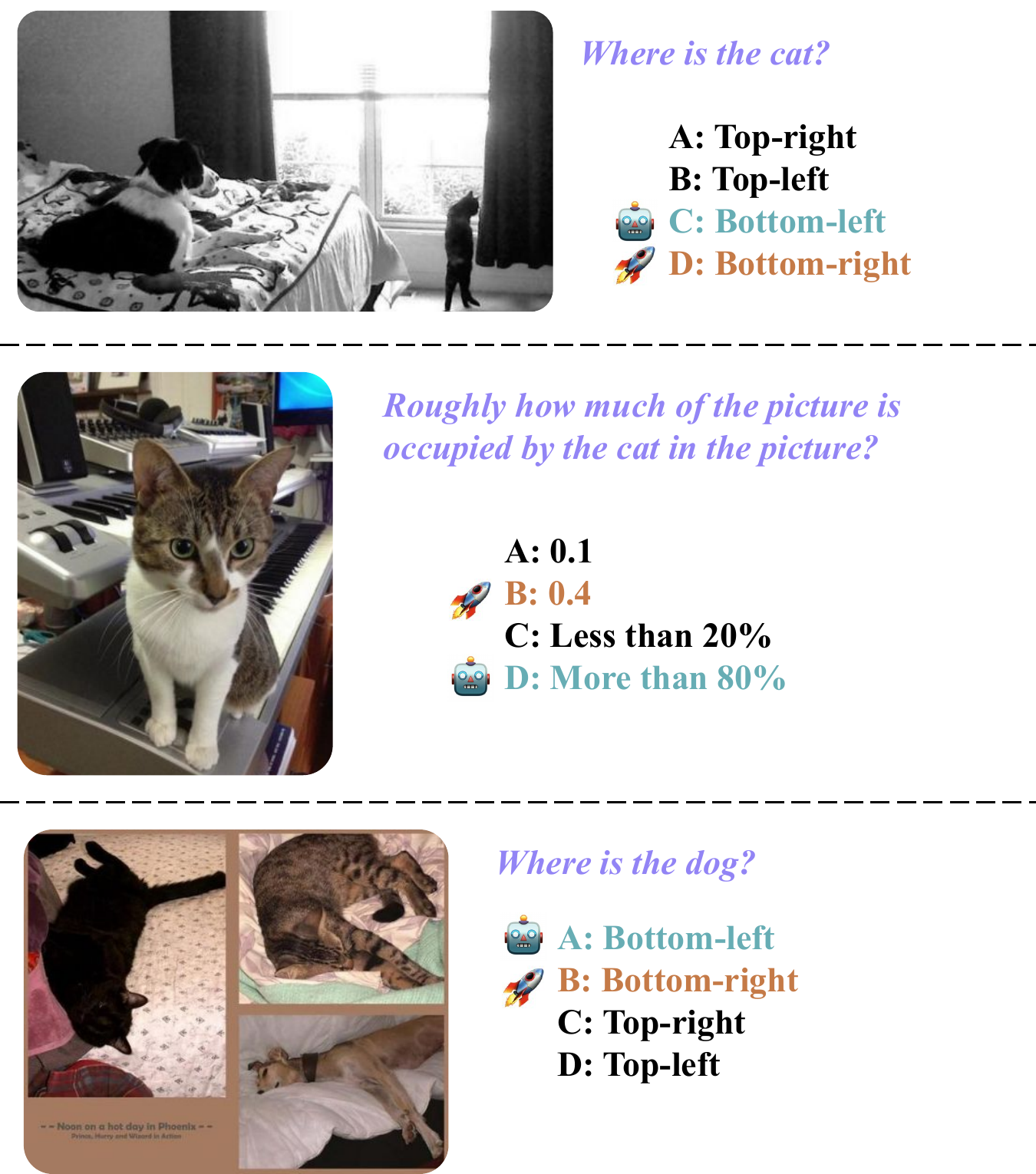}
      \caption{The visualizations of the \texttt{object localization} subtask from MMBench~\cite{liu2024mmbenchmultimodalmodelallaround}. Each subfigure shows the options selected by \textbf{\textcolor[HTML]{c57a47}{\ours}} and \textbf{\textcolor[HTML]{2596be}{CLIP}}~\cite{radford2021learningtransferablevisualmodels}. In every case, \ours selects the correct option, while CLIP does not.}
      \vspace{-8pt}
      \label{fig:MMBAppendixOne}
    \end{figure*}
}

\def\LanguageFLOPs#1{
    \begin{algorithm}[#1]
        \caption{\textsc{FLOPs of Language Transformers~\cite{vaswani2023attentionneed}}}
        \label{alg:alg1}
        \begin{algorithmic}[1]
            \Require
              \Statex \hspace*{\algorithmicindent}$n_{\text{ctx}}$ (average per-batch max caption token length), $n_{\text{vocab}}$ (vocab size), $d_{\text{model}}$ (model width),
              \Statex \hspace*{\algorithmicindent}$n_{\text{heads}}$ (attention head number), $d_{\text{key}}$ (key dimension), $d_{\text{ff}}$ (feed-forward width),
              \Statex \hspace*{\algorithmicindent}$n_{\text{layers}}$ (layer number).
            \Ensure  $F_{\text{total}}$
            
            \Statex \textbf{Embedding FLOPs}
            \State $F_{\text{emb}} \gets 2\,n_{\text{ctx}}\,n_{\text{vocab}}\,d_{\text{model}}$
            
            \Statex \textbf{Attention FLOPs (per layer)}
            \State $F_{qkv}      \gets 2\,n_{\text{ctx}}\,(3\,d_{\text{model}})\,(d_{\text{key}}\,n_{\text{heads}})$
            \State $F_{qk}       \gets 2\,n_{\text{ctx}}^{2}\,(d_{\text{key}}\,n_{\text{heads}})$
            \State $F_{\text{soft}} \gets 3\,n_{\text{heads}}\,n_{\text{ctx}}^{2}$
            \State $F_{\text{red}}  \gets 2\,n_{\text{ctx}}^{2}\,(d_{\text{key}}\,n_{\text{heads}})$
            \State $F_{\text{proj}} \gets 2\,n_{\text{ctx}}\,(d_{\text{key}}\,n_{\text{heads}})\,d_{\text{model}}$
            \State $F_{\text{attn}} \gets F_{qkv}+F_{qk}+F_{\text{soft}}+F_{\text{red}}+F_{\text{proj}}$
            
            \Statex \textbf{Feed-forward FLOPs (per layer)}
            \State $F_{\text{ff}} \gets 4\,n_{\text{ctx}}\,(d_{\text{model}}\,d_{\text{ff}})$
            
            \Statex \textbf{Total FLOPs}
            \State $F_{\text{total}} \gets F_{\text{emb}} + n_{\text{layers}}\,(F_{\text{attn}} + F_{\text{ff}})$
            \State \Return $3 \times F_{\text{total}}$
        \end{algorithmic}
    \end{algorithm}
}

\def\VisionFLOPs#1{
    \begin{algorithm}[#1]
        \caption{\textsc{FLOPs of Vision Transformers~\cite{dosovitskiy2021imageworth16x16words}}}
        \label{alg:alg2}
        \begin{algorithmic}[1]
            \Require                         
              \Statex \hspace*{\algorithmicindent}$n_{\text{patch}}$ (patch number), $d_{\text{patch}}$ (patch size), $n_{\text{channels}}$ (channel number), $d_{\text{model}}$ (model width),
              \Statex \hspace*{\algorithmicindent}$n_{\text{heads}}$ (attention head number), $d_{\text{key}}$ (key dimension), $d_{\text{ff}}$ (feed-forward width),
              \Statex \hspace*{\algorithmicindent}$n_{\text{layers}}$ (layer number).
            \Ensure  $F_{\text{total}}$
            
            \Statex \textbf{Embedding FLOPs}
            \State $F_{\text{emb}} \gets
                   2\,n_{\text{patch}}\,d_{\text{patch}}^{2}\,n_{\text{channels}}\,d_{\text{model}}$
            
            \Statex \textbf{Attention FLOPs (per layer)}
            \State $F_{qkv}      \gets 2\,n_{\text{patch}}\,(3\,d_{\text{model}})\,(d_{\text{key}}\,n_{\text{heads}})$
            \State $F_{qk}       \gets 2\,n_{\text{patch}}^{2}\,(d_{\text{key}}\,n_{\text{heads}})$
            \State $F_{\text{soft}} \gets 3\,n_{\text{heads}}\,n_{\text{patch}}^{2}$
            \State $F_{\text{red}}  \gets 2\,n_{\text{patch}}^{2}\,(d_{\text{key}}\,n_{\text{heads}})$
            \State $F_{\text{proj}} \gets 2\,n_{\text{patch}}\,(d_{\text{key}}\,n_{\text{heads}})\,d_{\text{model}}$
            \State $F_{\text{attn}} \gets F_{qkv}+F_{qk}+F_{\text{soft}}+F_{\text{red}}+F_{\text{proj}}$
            
            \Statex \textbf{Feed-forward FLOPs (per layer)}
            \State $F_{\text{ff}} \gets 4\,n_{\text{patch}}\,(d_{\text{model}}\,d_{\text{ff}})$
            
            \Statex \textbf{Total FLOPs}
            \State $F_{\text{total}} \gets F_{\text{emb}} + n_{\text{layers}}\, (F_{\text{attn}} + F_{\text{ff}})$
            \State \Return $3 \times F_{\text{total}}$
        \end{algorithmic}
    \end{algorithm}
}

\appendix

\section{Appendix}

\subsection{Limitations}
\label{sec:Limitations}
As discussed in the experiments section, \ours’s ability to capture compositional information is not yet complete. It shows relatively low accuracy on \texttt{swap object} and \texttt{swap attribute} compared to other SugarCrepe~\cite{hsieh2023sugarcrepe} tasks. We attribute this limitation to the fact that contrastive learning objectives still focus on aligning primarily lower-order statistics. Addressing this challenge requires exploring more refined information-theoretic measures for language-image alignment, which highlights a promising avenue for future research.

Also, due to computational constraints, we are unable to evaluate the scalability of \ours beyond 1.28 billion training samples. We acknowledge that CLIP~\cite{radford2021learningtransferablevisualmodels} and its variants may exhibit more favorable scaling behavior, as they jointly train both text and image encoders, whereas \ours keeps its text encoder frozen. Prior studies have shown that selectively unfreezing the last four layers of LLMs can substantially improve the scalability of image encoders without incurring heavy computational costs~\cite{chen2024internvlscalingvisionfoundation, stone2025learningvisualcompositionimproved}. How to efficiently fine-tune LLMs within mainstream language-image alignment pipelines remains an important direction for future work.

\subsection{More Visualizations from SugarCrepe~\cite{hsieh2023sugarcrepe}}
\label{sec:SCMoreVisualizations}
We present the visual results from evaluating \ours and CLIP~\cite{radford2021learningtransferablevisualmodels} on the seven compositional understanding tasks from SugarCrepe. For each caption, SugarCrepe generates a challenging negative caption by \texttt{add}, \texttt{replace}, or \texttt{swap} an \texttt{object}, \texttt{attribute}, or \texttt{relation} in the original caption. Models are asked to identify the correct caption based on caption-image cosine similarity. Both models use a ViT-B/16~\cite{dosovitskiy2021imageworth16x16words} backbone and are trained on 1.28B samples from DataComp-1B~\cite{gadre2023DataCompsearchgenerationmultimodal}. Each subfigure shows the captions selected by \ours (top) and CLIP (bottom). In every case, \ours selects the correct caption, while CLIP does not.

\ScAppendixOne{h}

\clearpage
\begin{figure*}[t]
  \centering

  \begin{subfigure}{\textwidth}
    \centering
    \includegraphics[width=0.99\linewidth]{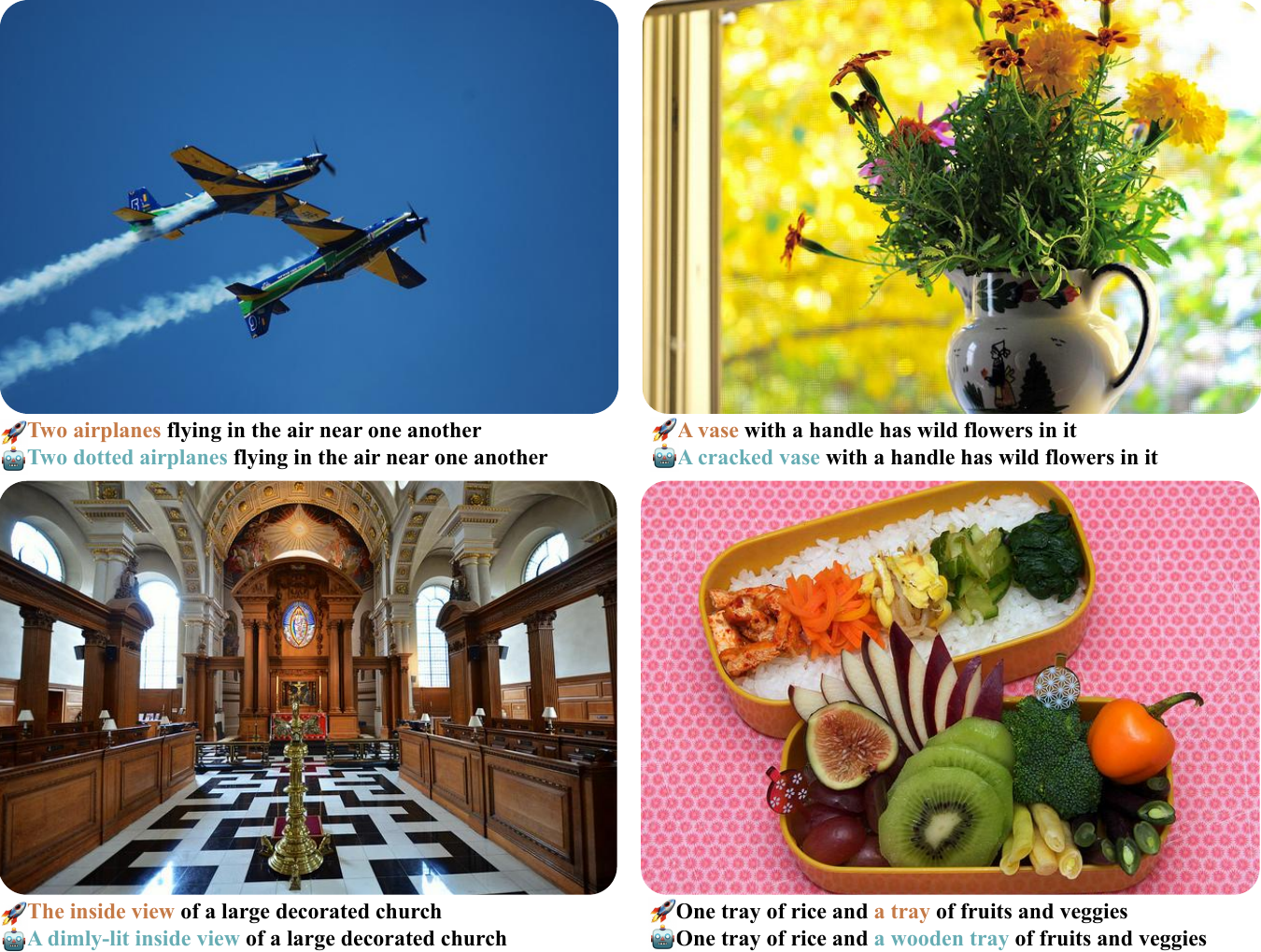}
    \caption{\texttt{Add Attribute}}
  \end{subfigure}

  \vspace{1em}

  \begin{subfigure}{\textwidth}
    \centering
    \includegraphics[width=0.99\linewidth]{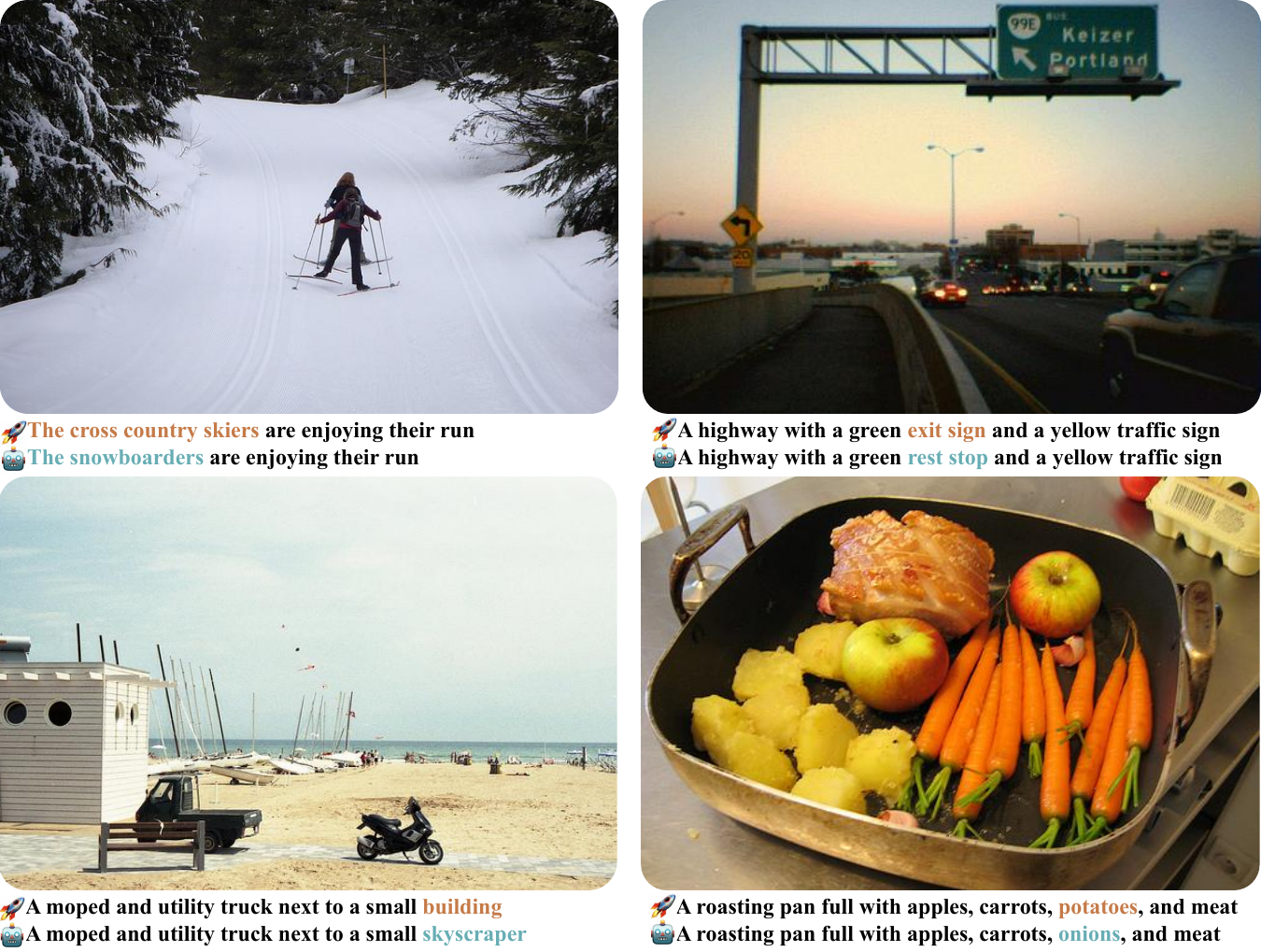}
    \caption{\texttt{Replace Object}}
  \end{subfigure}

  \caption{The visualizations of the \texttt{add attribute} and \texttt{replace object} tasks from SugarCrepe~\cite{hsieh2023sugarcrepe}. Each subfigure shows the captions selected by \ours (top) and CLIP~\cite{radford2021learningtransferablevisualmodels} (bottom). In every case, \ours selects the correct caption, while CLIP does not.}
\end{figure*}
\clearpage

\clearpage
\begin{figure*}[t]
  \centering

  \begin{subfigure}{\textwidth}
    \centering
    \includegraphics[width=0.99\linewidth]{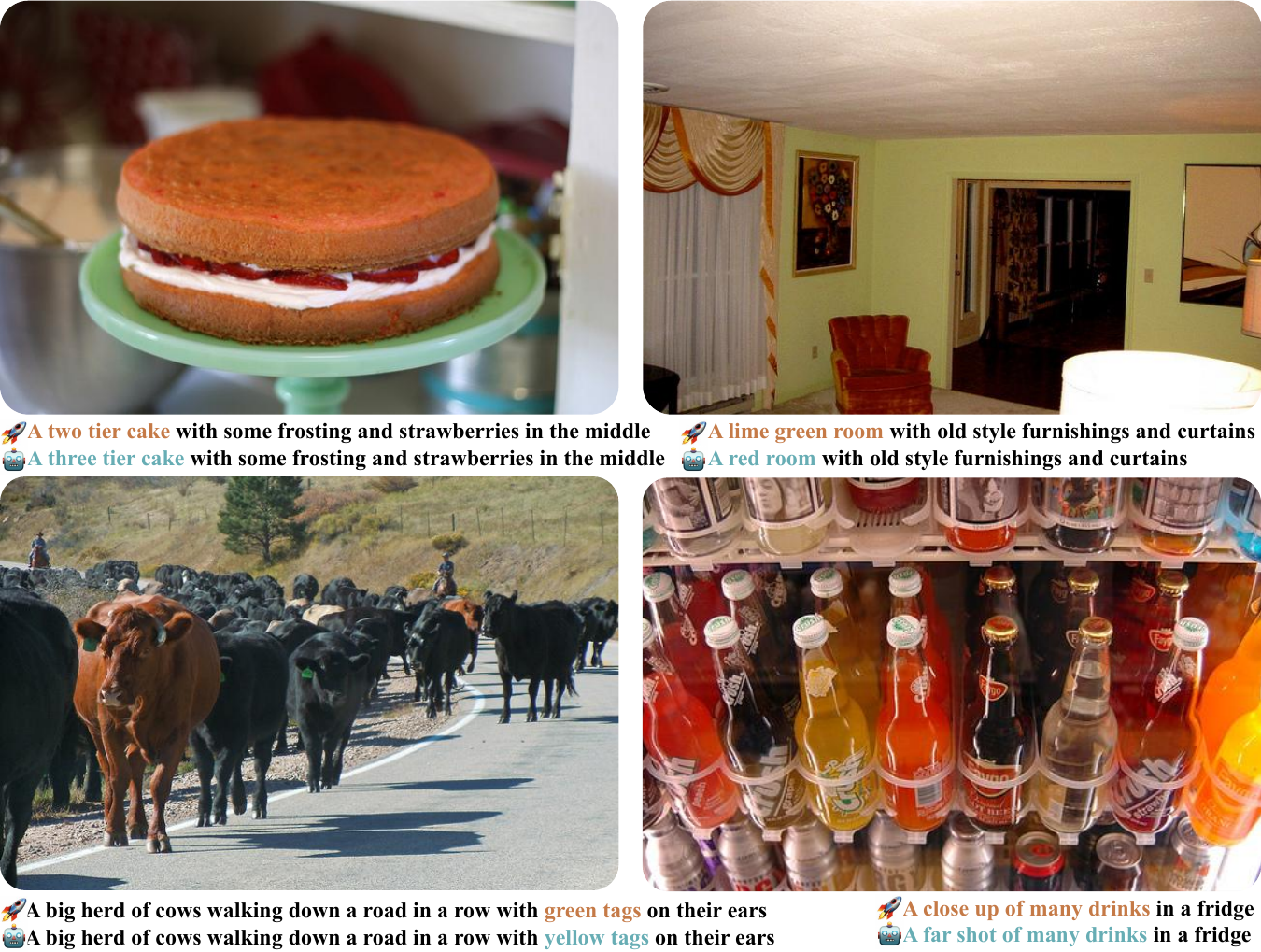}
    \caption{\texttt{Replace Attribute}}
  \end{subfigure}

  \vspace{1em}

  \begin{subfigure}{\textwidth}
    \centering
    \includegraphics[width=0.99\linewidth]{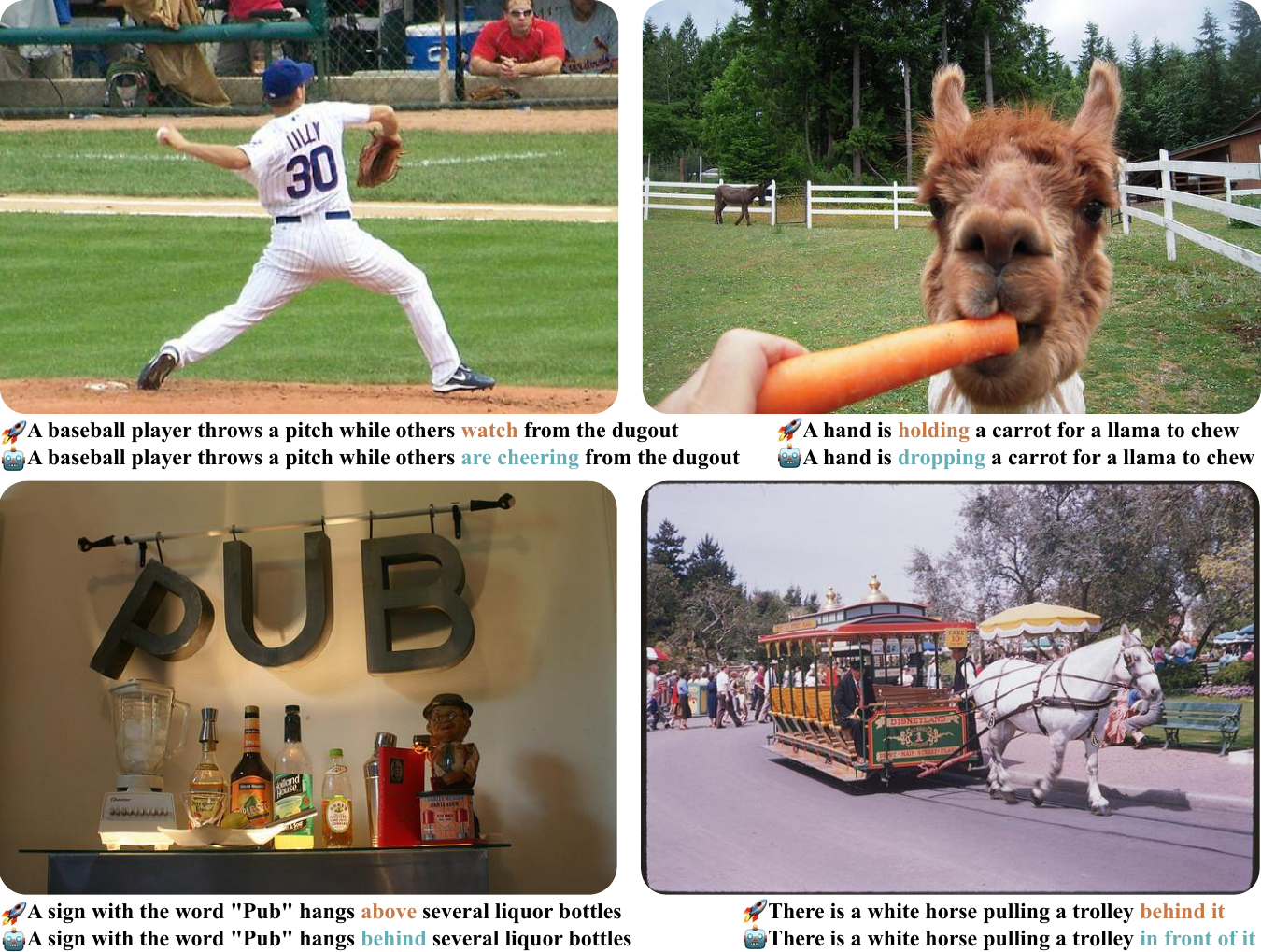}
    \caption{\texttt{Replace Relation}}
  \end{subfigure}

  \caption{The visualizations of the \texttt{replace attribute} and \texttt{replace relation} tasks from SugarCrepe~\cite{hsieh2023sugarcrepe}. Each subfigure shows the captions selected by \ours (top) and CLIP~\cite{radford2021learningtransferablevisualmodels} (bottom). In every case, \ours selects the correct caption, while CLIP does not.}
\end{figure*}
\clearpage

\clearpage
\begin{figure*}[t]
  \centering

  \begin{subfigure}{\textwidth}
    \centering
    \includegraphics[width=0.99\linewidth]{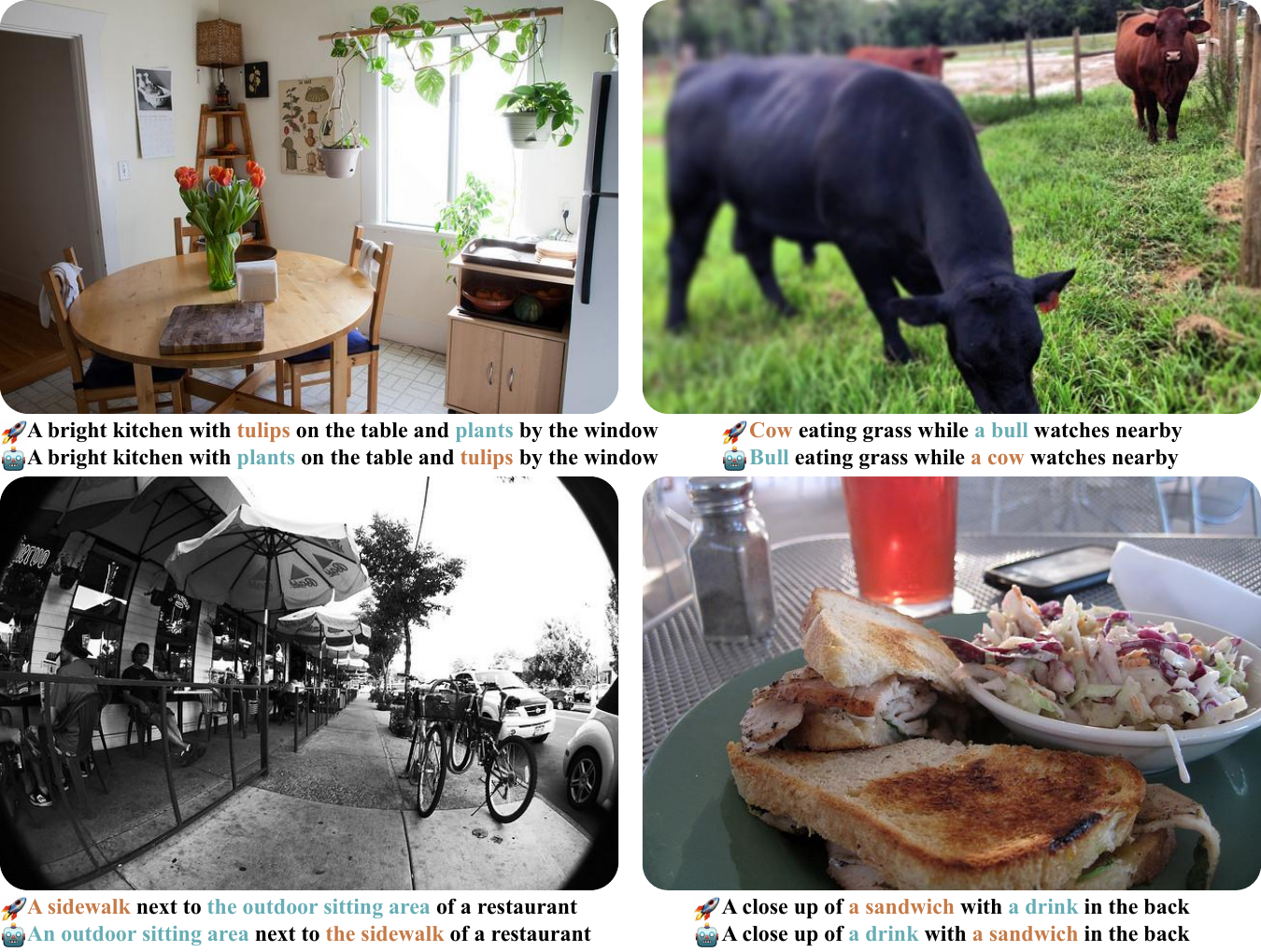}
    \caption{\texttt{Swap Object}}
    \end{subfigure}

  \vspace{1em}

  \begin{subfigure}{\textwidth}
    \centering
    \includegraphics[width=0.99\linewidth]{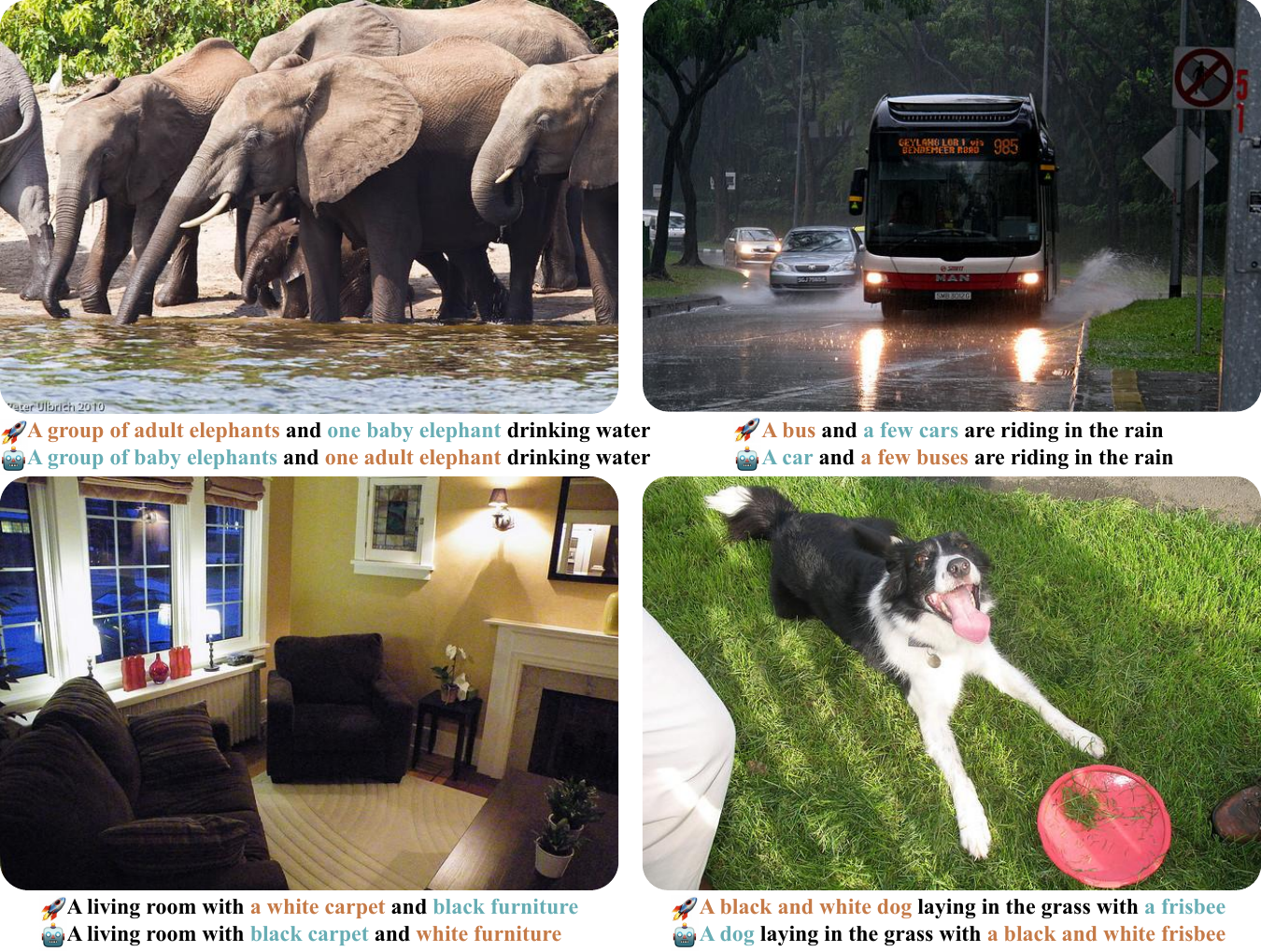}
    \caption{\texttt{Swap Attribute}}
  \end{subfigure}

  \caption{The visualizations of the \texttt{swap object} and \texttt{swap attribute} tasks from SugarCrepe~\cite{hsieh2023sugarcrepe}. Each subfigure shows the captions selected by \ours (top) and CLIP~\cite{radford2021learningtransferablevisualmodels} (bottom). In every case, \ours selects the correct caption, while CLIP does not.}
\end{figure*}
\clearpage

\subsection{More Visualizations from MMBench~\cite{liu2024mmbenchmultimodalmodelallaround}}
\label{sec:MMBenchTasksVisualizations}

We present the visual results from evaluating \ours and CLIP~\cite{radford2021learningtransferablevisualmodels} on five MMBench subtasks that specifically test models' compositional understanding. All questions are in the form of multiple-choice. Both models use a ViT-B/16~\cite{dosovitskiy2021imageworth16x16words} backbone and are trained on 1.28B samples from DataComp-1B~\cite{gadre2023DataCompsearchgenerationmultimodal}. Each subfigure shows the options selected by \textbf{\textcolor[HTML]{c57a47}{\ours (highlighted in orange)}} and \textbf{\textcolor[HTML]{2596be}{CLIP (highlighted in cyan)}}. In every case, \ours selects the correct option, while CLIP does not.

\MMBAppendixOne{h}

\clearpage
\begin{figure*}[t]
  \centering

  \begin{subfigure}{\textwidth}
    \centering
    \includegraphics[width=0.99\linewidth]{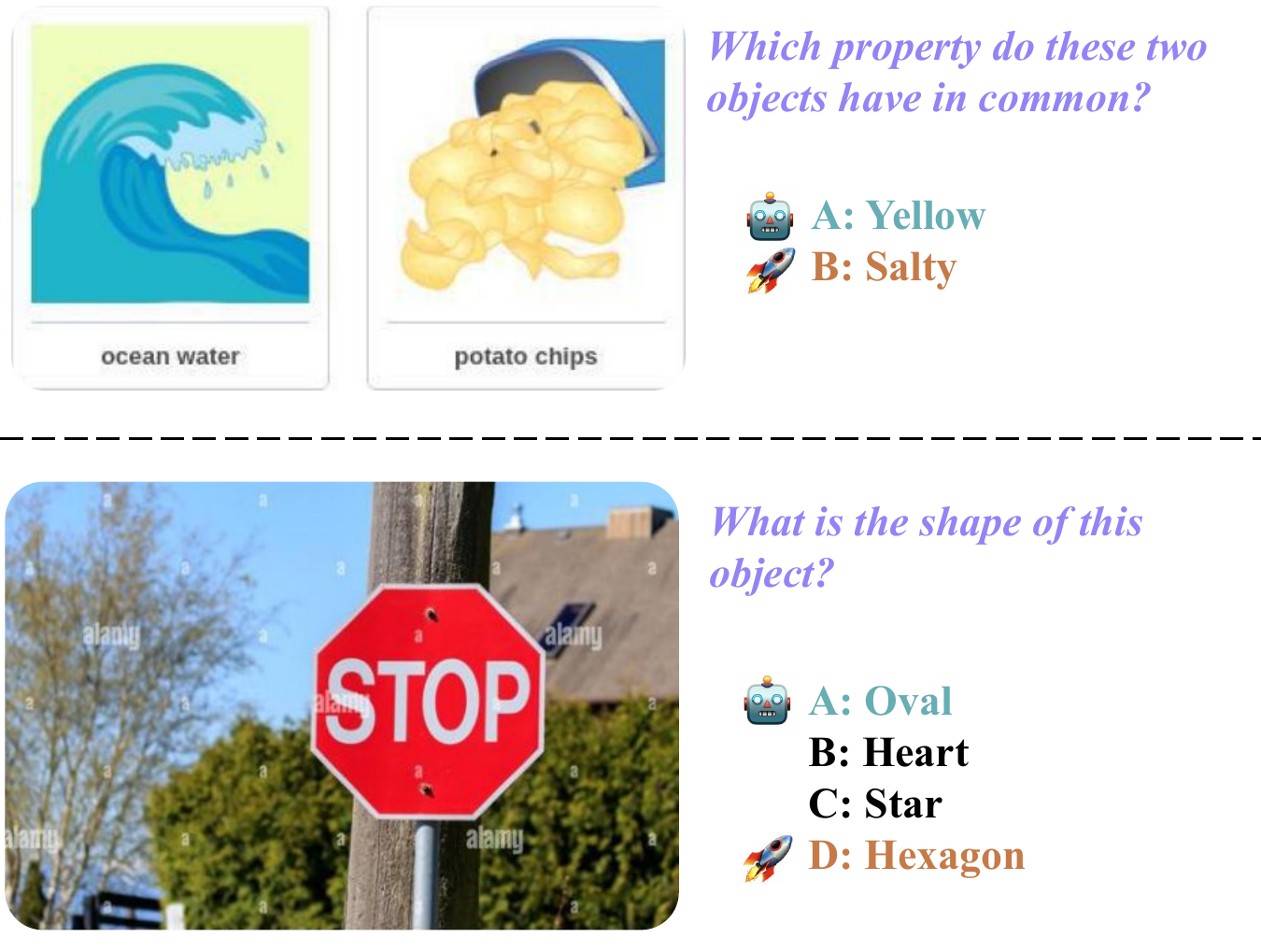}
    \caption{\texttt{Attribute Recognition}}
    \end{subfigure}

  \vspace{0.5em}

  \begin{subfigure}{\textwidth}
    \centering
    \includegraphics[width=0.99\linewidth]{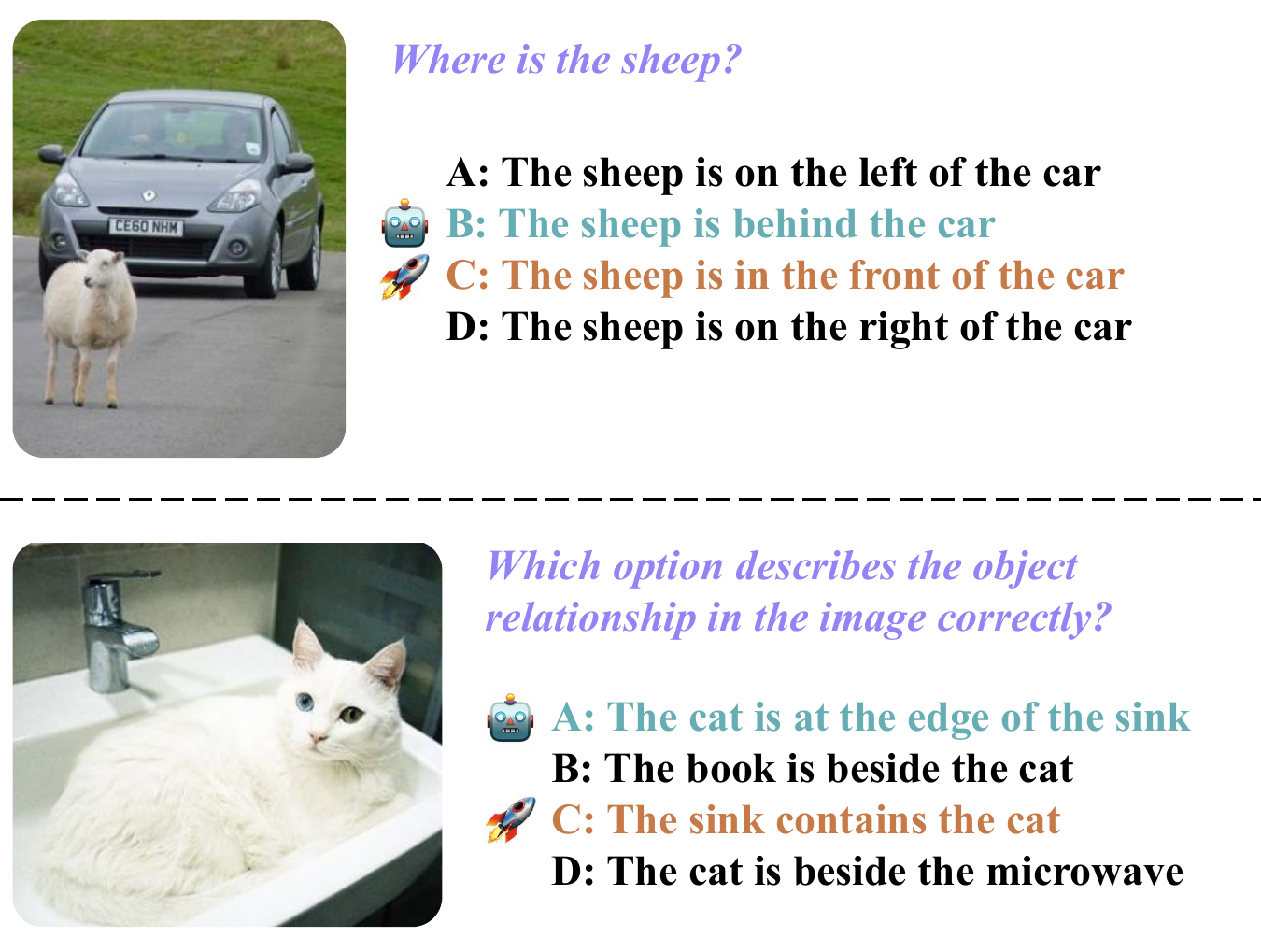}
    \caption{\texttt{Physical Relation}}
  \end{subfigure}

  \caption{The visualizations of the \texttt{attribute recognition} and \texttt{physical relation} subtasks from MMBench~\cite{liu2024mmbenchmultimodalmodelallaround}. Each subfigure shows the options selected by \textbf{\textcolor[HTML]{c57a47}{\ours}} and \textbf{\textcolor[HTML]{2596be}{CLIP}}~\cite{radford2021learningtransferablevisualmodels}. In every case, \ours selects the correct option, while CLIP does not.}
\end{figure*}
\clearpage

\clearpage
\begin{figure*}[t]
  \centering

  \begin{subfigure}{\textwidth}
    \centering
    \includegraphics[width=0.99\linewidth]{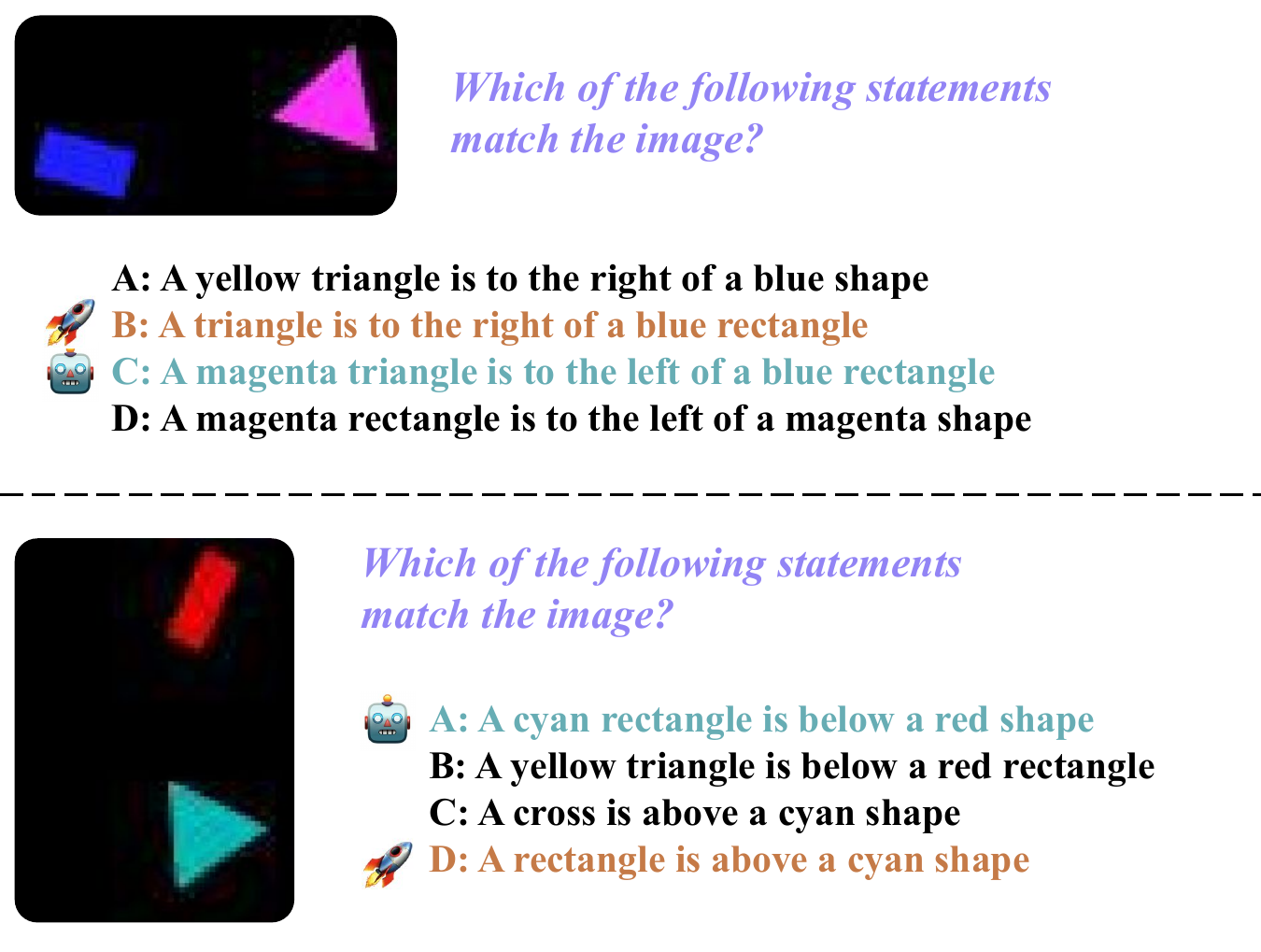}
    \caption{\texttt{Spatial Relation}}
    \end{subfigure}

  \vspace{0.5em}

  \begin{subfigure}{\textwidth}
    \centering
    \includegraphics[width=0.99\linewidth]{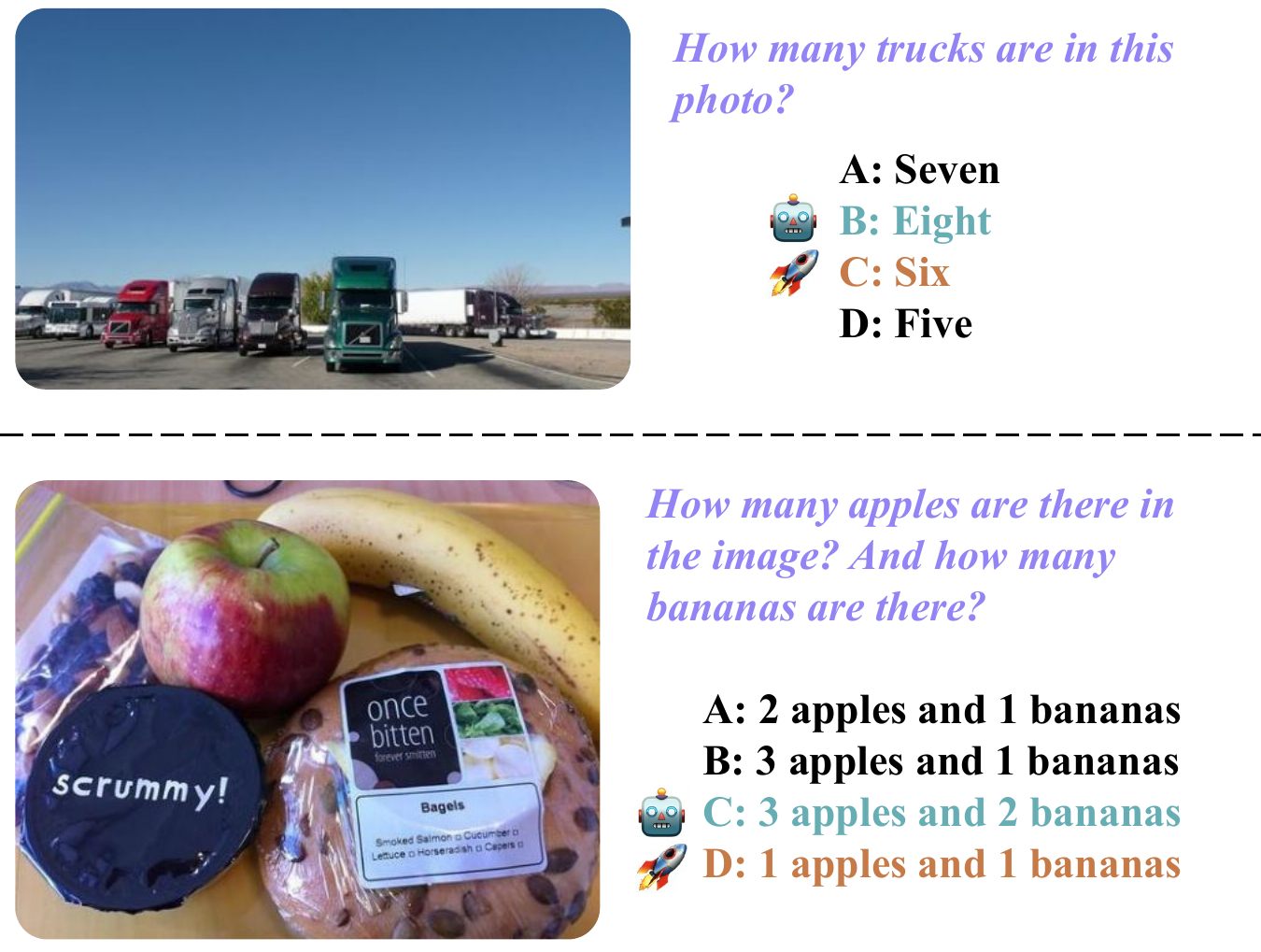}
    \caption{\texttt{Counting}}
  \end{subfigure}

  \caption{The visualizations of the \texttt{spatial relation} and \texttt{counting} subtasks from MMBench~\cite{liu2024mmbenchmultimodalmodelallaround}. Each subfigure shows the options selected by \textbf{\textcolor[HTML]{c57a47}{\ours}} and \textbf{\textcolor[HTML]{2596be}{CLIP}}~\cite{radford2021learningtransferablevisualmodels}. In every case, \ours selects the correct option, while CLIP does not.}
\end{figure*}
\clearpage

\subsection{The Calculation Details of FLOPs and Memory Usage}
\label{sec:FLOPsCal}
\LanguageFLOPs{h}
\VisionFLOPs{h}

\textbf{FLOPS}. Due to the truncation and padding applied by text tokenizers, the average per-sample FLOPs of each model should be calculated based on the average per-batch max caption token length. In our calculation, we approximate this value using the global max caption token length. Although this introduces some deviation from the exact values, our goal is not to report precise measurements but to demonstrate the difference in scaling behavior between the two models: CLIP~\cite{radford2021learningtransferablevisualmodels}’s FLOPs and memory footprint scale with $\mathcal{O}(n^2)$ complexity, while \ours achieves an amortized complexity of $\mathcal{O}(1)$.

In \cref{alg:alg1} and \cref{alg:alg2}, we present steps to calculate the per-sample FLOPs of the language~\cite{vaswani2023attentionneed} and vision transformers~\cite{dosovitskiy2021imageworth16x16words} used in this study. Both algorithms are adapted from \cite{hoffmann2022trainingcomputeoptimallargelanguage}. For CLIP, the per-sample FLOPs is approximated as the sum of the FLOPs from its language and vision transformers. Since \ours keeps its text encoder frozen, the per-sample FLOPs is estimated as that of its vision transformer alone. Throughout, we assume that a backward pass incurs twice the FLOPs of a forward pass.

\textbf{Memory Usage}. The memory usage of each model is estimated by monitoring GPU status using the \texttt{nvidia-smi} command. We observe that the exact memory consumption varies depending on factors such as training stages and GPUs. The reported results represent the average of five training runs conducted on different H800 GPUs with varying random seeds.

\subsection{Broader Impacts}
\label{sec:BroaderImpact}
Similar to other vision-language models (VLMs), \ours has significant potential for positive societal impacts. For instance, it advances multimodal understanding and can be fine-tuned to generate high-quality image captions for visually impaired individuals. As a weakly supervised method, \ours also reduces the reliance on heavily labeled datasets. However, \ours also carries potential negative societal impacts. It may inherit societal biases present in its training data, leading to harmful outcomes when deployed in sensitive applications. We advocate for cautious use of the model and recommend mitigating risks through careful dataset curation, bias analysis, and responsible deployment.

\subsection{Dataset Details}
\label{sec: DatasetInfo}
\subsubsection{Training Datasets}

\textbf{DataComp-1B~\cite{gadre2023DataCompsearchgenerationmultimodal}} is a large-scale, multimodal dataset consisting of approximately 1.4 billion text-image pairs curated from a massive pool of 12.8 billion web-crawled samples. The dataset is constructed using various filtering strategies --- including CLIP~\cite{radford2021learningtransferablevisualmodels} score filtering, text-based filtering, and image-based filtering --- to identify high-quality, semantically aligned pairs. Models trained on DataComp-1B have been shown to achieve notable gains on downstream tasks, including zero-shot classification on ImageNet-1K~\cite{5206848}.

\textbf{Recap-DataComp-1B~\cite{li2024recaptionbillionswebimages}}
builds upon DataComp-1B to address inherent noise and semantic misalignment in web-crawled captions. It fine-tunes a LLaVA-1.5~\cite{liu2024improvedbaselinesvisualinstruction} on LLaMA-3-8B~\cite{llama3modelcard} to recaption images with longer, richer, and more semantically aligned descriptions. Empirical studies indicate that Recap-DataComp-1B leads to gains in both cross-modal retrieval and text-to-image generation under complex queries.

\subsubsection{Evaluation Datasets}

\textbf{ImageNet~\cite{5206848}} is a large-scale, publicly available image dataset containing over 14 million images across more than 20,000 categories. For our zero-shot classification experiments, we use ImageNet-1K validation set, which consists of 50,000 images, with 50 images for each of the 1,000 classes. The input captions are constructed using the prompt template ``\texttt{It is a photo of \{label\}}.''

\textbf{COCO~\cite{lin2015microsoftcococommonobjects}} is a large-scale dataset for object detection, segmentation, keypoint detection, and image captioning, comprising 328,000 images. For our zero-shot retrieval experiments, we use \texttt{Val2017} split of 5,000 images and randomly select one of the five ground-truth captions for each image.

\textbf{Flickr30K~\cite{young-etal-2014-image}} consists of 31,000 images sourced from Flickr, each paired with five human-annotated reference captions. For our zero-shot retrieval experiments, we evaluate vision-language models on the test set of 1,000 images and randomly select one of the five ground-truth captions for each image.

\textbf{SugarCrepe~\cite{hsieh2023sugarcrepe}} is a benchmark designed to evaluate compositional understanding of vision-language models while addressing biases present in existing datasets. SugarCrepe employs large language models to generate fluent and challenging negative captions by \texttt{add}, \texttt{replace}, or \texttt{swap} an \texttt{object}, \texttt{attribute}, or \texttt{relation} in the original captions. Models are tasked with selecting the correct captions among these compositional distractors based on caption-image cosine similarity.

\textbf{MMBench~\cite{liu2024mmbenchmultimodalmodelallaround}} is a benchmark designed to evaluate vision-language models across a broad range of perception and reasoning abilities. It features a diverse set of evaluation questions curated with strict quality control and introduces a CircularEval strategy that uses large language models to map free-form outputs to multiple-choice answers. It supports bilingual evaluation in English and Chinese.

\textbf{MME~\cite{fu2024mmecomprehensiveevaluationbenchmark}}
is a benchmark designed to evaluate vision-language models across 14 subtasks covering both perception and cognitive abilities. It provides systematic evaluations using manually crafted instruction-answer pairs. MME's standardized instructions enable fair comparisons between models without relying on prompt engineering.

\textbf{POPE~\cite{li2023evaluatingobjecthallucinationlarge}}
is a benchmark designed to systematically assess object hallucination in vision-language models, a common issue in which models tend to generate objects that are inconsistent with the target images. It innovates a polling-based query method that offers a more stable and flexible evaluation of hallucinated content.

\textbf{ScienceQA~\cite{lu2022learnexplainmultimodalreasoning}}
is a vision-language model benchmark comprising about 21,000 multiple-choice science questions. Each question is accompanied by rich annotations that encourage models to generate chain-of-thought~\cite{wei2023chainofthoughtpromptingelicitsreasoning} (CoT) responses and enable the study of multi-hop reasoning. ScienceQA supports evaluation across textual, visual, and diagrammatic modalities.

\textbf{TextVQA \cite{singh2019vqamodelsread}} is a benchmark designed to assess vision-language models’ ability to recognize text within images. It consists of 45,336 questions across 28,408 images, where successful answering requires models to accurately read text in the image and reason about it in the context of both the image and the question.